\documentclass[11pt, a4paper, logo, copyright]{style}

\pdftrailerid{redacted}
\makeatletter
\renewcommand\bibentry[1]{\nocite{#1}{\frenchspacing\@nameuse{BR@r@#1\@extra@b@citeb}}}
\makeatother
\usepackage{color}
\usepackage{rotating} 
\usepackage{tabularray}
\usepackage{kantlipsum, lipsum}
\usepackage{dsfont}
\usepackage{gdm-colors}
\usepackage[utf8]{inputenc}   
\usepackage{newunicodechar}  
\usepackage{amssymb}          

\usepackage{wasysym,marvosym}
\usepackage{ulem}
\usepackage{caption}
\usepackage{floatrow}
\usepackage{tikz}
\usetikzlibrary{shapes.geometric, arrows.meta, positioning}
\usepackage{amsmath}
\usepackage{algorithm}
\usepackage{algpseudocode}
\usepackage{listings}
\usepackage{enumitem} 
\usepackage{lipsum}
\usepackage[most]{tcolorbox}
\definecolor{thinkcolor}{RGB}{227,196,144}
\definecolor{observecolor}{RGB}{153,201,227}
\definecolor{explorecolor}{RGB}{178,217,200}

\definecolor{GreenHaze}{rgb}{0,0.564,0.333}
\definecolor{Jade}{rgb}{0,0.729,0.45}
\definecolor{Shamrock}{rgb}{0.243,0.847,0.596}
\definecolor{SunsetOrange}{rgb}{1,0.29,0.278}
\definecolor{Salmon}{rgb}{1,0.486,0.478}
\definecolor{AlizarinCrimson}{rgb}{0.831,0.117,0.129}

\tcbset{
    common/.style={
		enhanced,
		arc=0mm,
		fonttitle=\large\bfseries,
		coltitle=black,
		attach boxed title to top left={xshift=0mm,
										yshift=-0.50mm},
		boxed title style={
			skin=enhancedfirst jigsaw,
			size=small,
			arc=5mm,
			bottom=0mm,
			left=8mm,
			right=18mm,
			top=1mm},
			boxrule=0pt,
			frame hidden},
    thinkstyle/.style={
		common,
		colbacktitle=thinkcolor,
		colframe=thinkcolor,
		colback=thinkcolor!40,
		borderline north={4pt}{0pt}{thinkcolor}},
	observestyle/.style={
		common,
		colbacktitle=observecolor,
		colframe=observecolor,
		colback=observecolor!40,
		borderline north={4pt}{0pt}{observecolor}}
}

\newtcolorbox{think}{thinkstyle,title=Prompt Template}
\newtcolorbox{observe}{observestyle,title=observe}
\newtcolorbox{custom}[2][gray]{
	common,
	title=#2,
	colbacktitle=#1,
	colframe=#1,
	colback=#1!40,
	borderline north={4pt}{0pt}{#1}}
\usepackage{array, multirow, tabularx, booktabs, makecell}

\definecolor{interest_colframe}{rgb}{0.8, 0.878, 0.871}
\definecolor{interest_colback}{rgb}{0.918, 0.953, 0.949}

\definecolor{tag_colframe}{rgb}{0.965, 0.898, 0.847}
\definecolor{tag_colback}{rgb}{0.988, 0.961, 0.941}

\definecolor{exp_colback}{rgb}{0.949, 0.965, 0.980}
\definecolor{exp_colframe}{rgb}{0.878, 0.922, 0.965}

\tcbset{
    promptbox/.code args={#1/#2}{
        \tcbset{
            enhanced,
            arc=0mm,
            colframe=#1, 
            colback=#2, 
            coltitle=black,
            fonttitle=\large\bfseries,
            attach boxed title to top left={xshift=0mm, yshift=-1.0mm},
            boxed title style={
                skin=enhancedfirst jigsaw,
                size=small,
                arc=3mm,
                bottom=0mm,
                left=8mm,
                right=8mm,
                top=1mm,
                colback=#1
            },
            boxrule=0pt,
            frame hidden,
            borderline north={4pt}{0pt}{#1},
        }
    }
}

\newcommand{\assignmentQuestionName}{Question}

\usepackage{booktabs}
\usepackage{arydshln}
\usepackage{dashrule}

\usepackage[authoryear, sort&compress, round]{natbib}

\usepackage{bbding}
\usepackage[T1]{fontenc}   
\usepackage{hyperref}       
\usepackage{url}           
\usepackage{booktabs}       
\usepackage{nicefrac}       
\usepackage{microtype}      
\usepackage{amsmath}
\usepackage{graphicx}
\usepackage{multicol}
\usepackage[nameinlink]{cleveref}
\usepackage{bbm}
\usepackage{multirow}
\usepackage{soul}
\usepackage{float}
\usepackage{wrapfig}
\usepackage{blindtext}
\usepackage{tablefootnote}
\usepackage{amsfonts}
\usepackage[flushleft]{threeparttable}
\usepackage{colortbl}
\usepackage{mathtools,amssymb}
\usepackage{bm}
\usepackage{makecell}
\usepackage{caption}
\usepackage{capt-of}
\usepackage{array}
\usepackage{calc}      
\usepackage{subcaption}  
\usepackage[bottom]{footmisc}
\usepackage{fontawesome}
\usepackage{tabularx}

\graphicspath{{Figures/}}

\title{
\Large \bfseries Higher Satisfaction, Lower Cost:
\Large \bfseries A Technical Report on How LLMs Revolutionize Meituan's Intelligent Interaction Systems
}

\renewcommand{\today}{}
\author{\large LongCat Interaction Team}

\begin{abstract}
Enhancing customer experience is essential for business success, particularly as service demands grow in scale and complexity. Generative artificial intelligence and Large Language Models~(LLMs) have empowered intelligent interaction systems to deliver efficient, personalized, and 24/7 support. In practice, intelligent interaction systems encounter several challenges: (1) Constructing high-quality data for cold-start training is difficult, hindering self-evolution and raising labor costs. (2) Multi-turn dialogue performance remains suboptimal due to inadequate intent understanding, rule compliance, and solution extraction. (3) Frequent evolution of business rules affects system operability and transferability, constraining low-cost expansion and adaptability. (4) Reliance on a single LLM is insufficient in complex scenarios, where the absence of multi-agent frameworks and effective collaboration undermines process completeness and service quality. (5) The open-domain nature of multi-turn dialogues, lacking unified golden answers, hampers quantitative evaluation and continuous optimization. 

\hspace{1.3em} To address these challenges, we introduce WOWService, an intelligent interaction system tailored for industrial applications. With the integration of LLMs and multi-agent architectures, WOWService enables autonomous task management and collaborative problem-solving. Specifically, WOWService focuses on core modules including data construction, general capability enhancement, business scenario adaptation, multi-agent coordination, and automated evaluation. Currently, WOWService is deployed on the Meituan App, achieving significant gains in key metrics, e.g., User Satisfaction Metric 1~(USM 1) -27.53\% and User Satisfaction Metric 2~(USM 2) +25.51\%, demonstrating its effectiveness in capturing user needs and advancing personalized service.
\end{abstract}

\begin{document}
\maketitle
\begin{figure}[h!]
\centering
\includegraphics[width=1\textwidth]{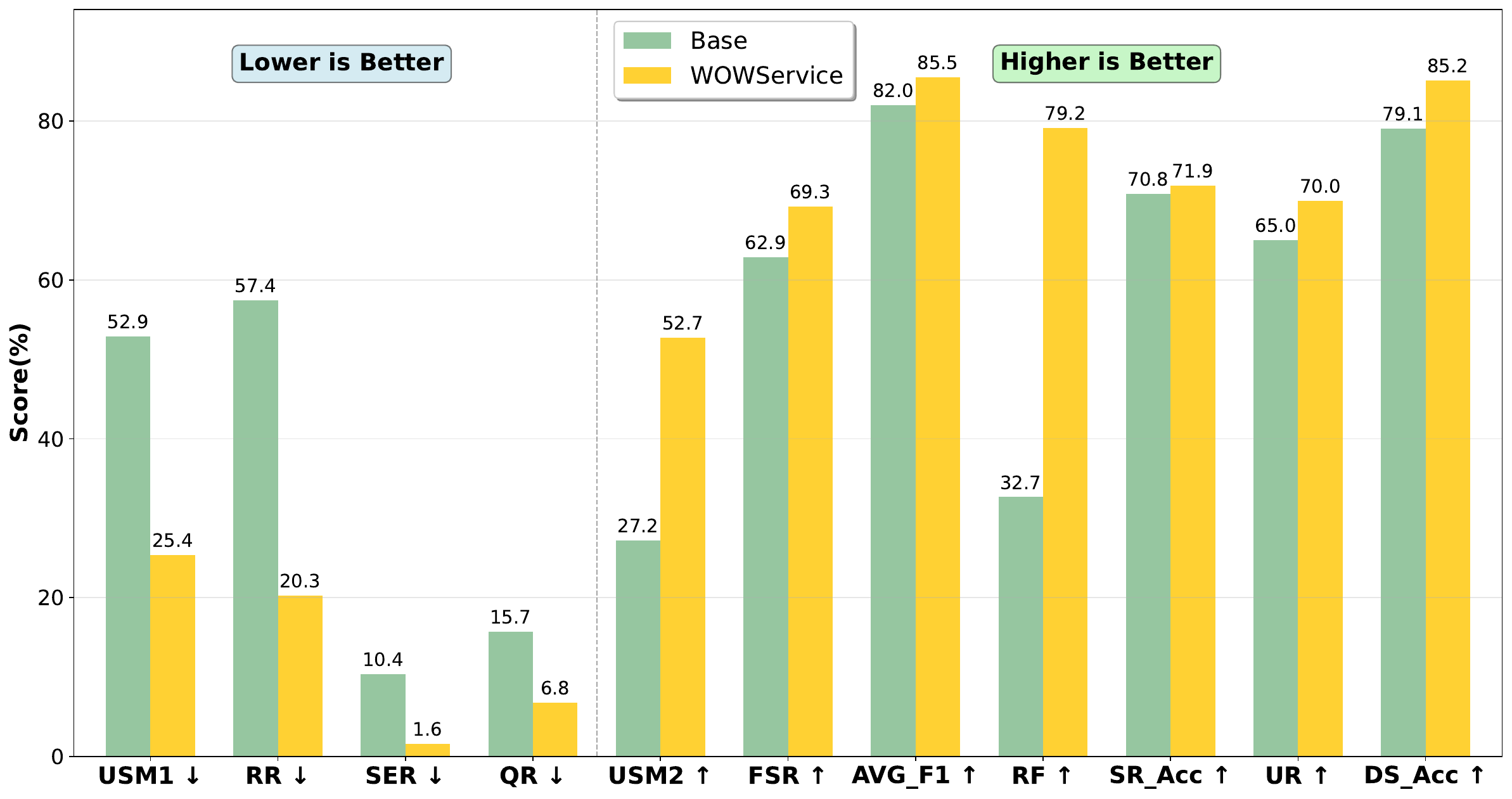}
\vspace{-1em}
\caption{The comparative results of WOWService and the base model across 11 key metrics, which demonstrate that WOWService achieves notable superiority in business applications.}

\label{fig:abs}
\end{figure}
\newpage
\setcounter{tocdepth}{2} 

\tableofcontents

\newpage
\section{Introduction}
In commercial environments, improving the dialogue interaction experience has become a critical factor that influences business performance~\citep{mccoll2019gaining}. As enterprises continue to expand, interaction service demands have become more diversified, round-the-clock, and large-scale. Traditional human-operated interaction service often faces numerous challenges when addressing massive volumes of user requests and complex application scenarios, including high operational costs, insufficient standardized management, and inconsistent service experiences~\citep{vijayakumar2023transforming}. Benefiting from clearly defined application boundaries, extensive historical data accumulation, and highly standardized business processes, the interaction sector has emerged as an ideal scenario for the application of generative artificial intelligence technologies~\citep{inavolu2024exploring}. Driven by this trend, enterprises have increasingly adopted intelligent interaction systems that integrate advanced artificial intelligence technologies to efficiently handle large volumes of standardized inquiries, provide uninterrupted 24/7 service~\citep{desmal2023automated}, and effectively free up human resources from repetitive and tedious tasks, thus fostering an intelligent and collaborative ecosystem between humans and machines~\citep{qian2024societal}. Intelligent interaction systems not only achieve Pareto optimization by controlling operational costs while enhancing customer experience and business efficiency~\citep{fawcett1883manual}, but also provide robust technical support for business expansion.

In the context of local lifestyle services, dialogue interactions exhibit significant periodic fluctuations and high concurrency characteristics~\citep{shankar2022online, bao2022customers}. Additionally, user demands are often expressed with semantic ambiguity, and problem descriptions tend to be highly real-time and dynamic. The interaction process may also involve multiple parties, such as users, merchants, and delivery riders, further increasing the complexity of interaction scenarios and posing greater challenges to intelligent interaction systems~\citep{pi2024contact}. The unique nature of local lifestyle services is reflected not only in the diversity and complexity of user needs but also in the expectation of users to receive immediate and accurate solutions~\citep{zhou2020design, carrera2022context}. In such multi-party collaborative scenarios, the system must be able to flexibly dispatch and coordinate resources among all involved parties, ensuring efficient information flow.

Conventional models based on the Transformer architecture, such as BERT, efficiently describe global dependencies in text through attention mechanisms, enhancing the capabilities of understanding natural languages~\citep{vaswani2017attention,devlin2019bert}.  This enables automated question-and-answering (Q\&A) systems to better understand customer needs, thereby improving the accuracy and user experience of intelligent interaction systems. With in-depth research on model scaling laws, model sizes have continuously expanded, driving the emergence of large-scale models such as ChatGPT~\citep{achiam2023gpt} and DeepSeek~\citep{liu2024deepseek}. These large models have further improved intelligent interaction systems to understand complex contexts, enabling efficient and personalized interactions~\citep{brown2020language,adam2021ai}. Meanwhile, the continuous improvement of artificial intelligence capabilities has prompted intelligent interaction systems to introduce agent architectures. Agents not only understand and respond to user requests but also possess autonomous task management and process automation execution capabilities, achieving a transformation from ``conversational interaction service'' to ``task-oriented interaction service''~\citep{zhang2024emotional}. To address increasingly complex and diverse interaction needs, intelligent interaction systems are gradually advancing toward the multi-agent stage. Multi-agent systems enable collaboration and division of labor among different agents, working together to handle complex service requests across departments, systems, and scenarios, such as Alibaba's AliMe~\citep{QiuLWGCZCHC17}, JingDong JIMI~\citep{JingDongJIMI}, and Ant Intelligent Interaction System~\citep{AntICS}. This collaborative mechanism significantly improves the flexibility, scalability, and depth of service of intelligent interaction systems.

Recent advancements in Large Language Models (LLMs) have significantly enhanced their abilities in language understanding and content generation, especially for intelligent interaction scenarios~\citep{li2025beyond,bai2024mt}. Although increasing efforts focus on leveraging LLMs to enhance the performance of intelligent interaction systems, most studies remain limited to small-scale, offline benchmarks, restricting real-world applicability~\citep{kwan2024mt,zhang2025turnbench}. Thus, effectively integrating LLMs into large-scale industrial scenarios—to accurately understand user intent and deliver diverse services—remains largely unaddressed. In practical business applications, current intelligent interaction systems face several challenges: (1) During the cold-start training phase, constructing high-quality data is difficult. Mainstream training methods rely on extensive manual labeling~\citep{team2025longcat}, which hinders self-evolution and increases labor costs. (2) In multi-turn dialogues, there is a notable space for improvement in understanding user intent, adhering to service rules, and extracting solutions~\citep{castillo2024beyond}. Model generalization and the complexity of business scenarios often result in suboptimal dialogue performance, making it hard to meet personalized needs. (3) As business rules and Q\&A content evolve, systems struggle with operability, transferability, and self-evolution, limiting low-cost business expansion and adaptability~\citep{fang2025comprehensive}. (4) Most systems rely on a single LLM, but complex real-world scenarios often exceed any individual model’s capabilities~\citep{zhang2024large}. The absence of end-to-end multi-agent frameworks and efficient collaboration restricts intelligent process completeness and model cooperation, affecting overall service quality. (5) Multi-turn dialogues are open-domain, lacking unified golden answers~\citep{guan2025evaluating}, which complicates automated evaluation and hinders quantitative performance assessment and continuous system optimization.

To this end, we propose WOWService, a comprehensive intelligent interaction framework tailored for industrial applications, including data construction, general capability enhancement, business scenario adaptation, multi-agent coordination, and automated evaluation. It continuously strengthens LLMs’ domain skills, enabling proactive and actionable service for diverse interaction scenarios. Specifically, we implement a robust training pipeline comprising four stages: Continual Pre-Training~(CPT), Supervised Fine-Tuning~(SFT), Direct Preference Optimization~(DPO), and Reinforcement Learning~(RL), to boost model capabilities in practice. Moreover, as business complexity grows, single LLMs are insufficient for service demands. WOWService evolves from a single-agent to a multi-agent architecture, introducing specialized agents that collaborate on targeted business demands. For automated evaluation, WOWService features a full-process system, from base model performance to end-to-end agent effectiveness, as a basis for model iteration in open-domain tasks. Currently, WOWService has been successfully deployed on the Meituan App, delivering notable improvements in key metrics, e.g., User Satisfaction Metric 1~(USM 1) -27.53\% and User Satisfaction Metric 2~(USM 2) +25.51\%, clearly demonstrating its value in understanding user needs and enhancing personalized service. Our main contributions can be summarized as follows:
\begin{itemize}
\item WOWService uses a dual-driven strategy, combining data-driven and knowledge-driven approaches, to build and refine high-quality training data. Through self-refinement training enhancement, our framework ensures robust performance in dynamic interaction scenarios.
\item We introduce a multi-stage training pipeline to strengthen competence in business-specific applications, including CPT, SFT, DPO, and RL stages. This progressive approach significantly improves the accuracy and targeting of the models’ interactions.
\item Our framework features a scalable mechanism that extends service coverage, enabling rapid adaptation to evolving business needs and seamless integration of new domains.
\item Through a multi-agent architecture that serves diverse business applications, our framework enables proactive and actionable service, facilitating collaborative problem-solving and ensuring that customers receive timely, context-aware assistance tailored to their specific needs.
\item We establish a thorough evaluation framework to systematically assess performance across various metrics, which ensures continuous improvement and maintains high service quality.
\end{itemize}

\section{Model Section}
\label{sec:RecGPT_Workflow}

\subsection{Continual Pre-Training Stage}
In dialogue interaction contexts, LLMs must demonstrate domain expertise to deliver appropriate solutions~\citep{DBLP:journals/mlc/LiWXWLCLFDHSL25}. However, enhancing domain-specific capabilities solely through post-training methods has inherent limitations, so we adopt continual pre-training~\citep{DBLP:conf/iclr/KeSLKK023} and incorporate large-scale dialogue interaction data to build a more robust foundation model.

Continual pre-training for the dialogue interaction domain faces two primary challenges: first, achieving a balance between general and dialogue interaction-specific capabilities; and second, constructing refined and logically-complete domain data. To address these challenges, we adopt a multifaceted approach that enhances domain-specific performance while preserving or even improving general capabilities, as detailed in the following subsections.

\subsubsection{Balance between General and Domain Capabilities}
During the construction of domain-specific models, continual pre-training in industry-specific domains often struggles with the core issue of degradation in general capabilities, which can lead to shortcomings in essential skills such as instruction following and reasoning required for dialogue interaction LLM applications. This degradation in general capabilities primarily arises from:
\begin{itemize}
\item \textbf{Catastrophic Forgetting:} When domain-specific data is layered onto the base model for training, the original knowledge representations are prone to being overwritten~\citep{huang2024mitigating}.
\item \textbf{Data Distribution Shift:} In scenarios involving mixed general and domain data, the absence of fine-grained distribution information from the original pre-training corpus, coupled with variations in data quality, can adversely impact original general capabilities.
\end{itemize}

For instance, in practical continual pre-training experiments on the Yi-34B~\citep{DBLP:journals/corr/abs-2403-04652} model for the dialogue interaction domain, we observe a 2.93\% decline in general capabilities even when strictly adhering to the proportions of general data injection as prescribed in the official technical report, following the completion of training.

To enhance dialogue interaction capabilities while preserving---or even improving---the base model's general capabilities, we curate high-quality general data incorporated into training and implement the method of adaptive data mixture optimization. Data mixtures play a pivotal role in influencing model performance during large-scale pre-training~\citep{held2025optimizing, linwood2024optimizing}. However, conventional data mixture experimentation is often hindered by excessive resource demands and extended experimental timelines. Drawing inspiration from scaling laws for data mixtures in cold-start pre-training~\citep{liuregmix}, we innovatively extend these principles to the continual pre-training phase. Specifically, we employ small-scale proxy models to rapidly identify optimal data mixtures, which are then scaled to full model training, enabling the determination of the optimal mixtures in a single iteration.

\begin{figure}[htbp]
\centering
\includegraphics[width=0.8\textwidth]{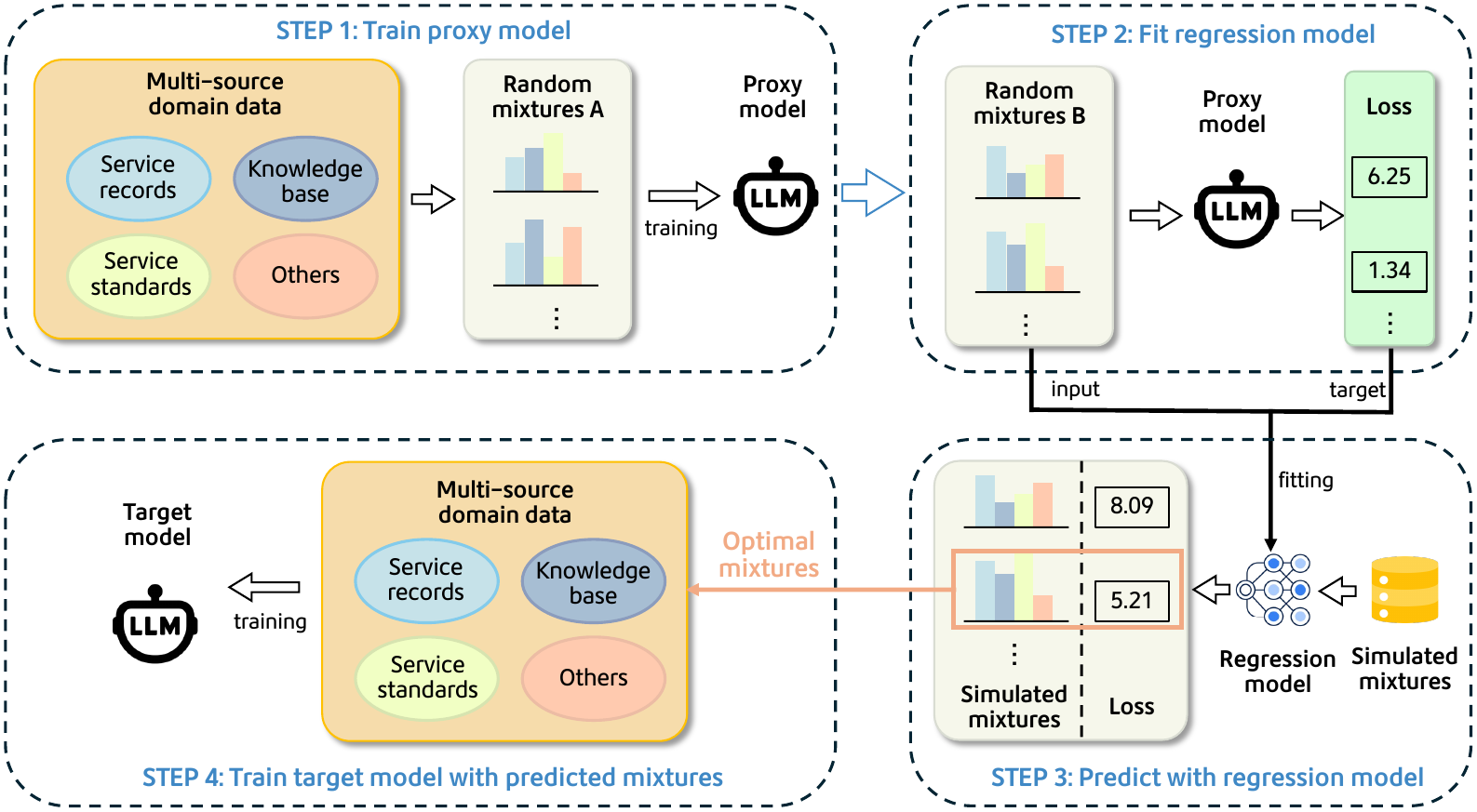}
\caption{Diagram of adaptive data mixtures optimization framework in continual pre-training.}
\vspace{-0.2cm}
\label{fig:auto_mixtures}
\end{figure}

The core steps of this method are as follows (as shown in Figure~\ref{fig:auto_mixtures}):
    \begin{enumerate}
    \item Generate random mixtures of data proportions based on multi-source domain data and general data; train small-scale proxy models on these mixtures according to the specified ratios.
    \item Treat the data ratio values as multi-dimensional feature vectors and use the validation set loss from the trained proxy model as labels to fit a regression model.
    \item Simulate an expanded space of larger-scale data mixtures and leverage the regression model to predict the optimal target values, thereby deriving the best mixture ratio.
    \item Sample data using the simulated optimal mixture ratio and train the target model accordingly.
    \end{enumerate}

In the training of LongCat, this method yields performance improvements compared to manually tuned optimal ratios. Moreover, the method shows over 75\% greater efficiency than manual tuning.
\vspace{-0.2cm}

The proportion of general data to domain-specific data has a significant impact on model performance. Experimental results indicate that increasing the proportion of domain-specific data leads to a certain decline in general capabilities. However, at an optimal data ratio—approximately 80\% general data—a favorable trade-off is achieved, where the decline in general capabilities is minimal while domain-specific capabilities are significantly enhanced.

\subsubsection{Construction of Refined Domain-Specific Data}

High-quality domain-specific data is critical for our models, but the raw data from numerous heterogeneous sources suffers from low knowledge density, inconsistent quality, and high redundancy. To address this, we build a sophisticated data processing pipeline to refine this data. Beyond standard filtering, we employ like \textit{quality filtering} using strong models, and \textit{dialogue chain-of-thought rewriting} to augment reasoning. This process yields tens of billions of high-quality tokens for training.

Furthermore, given the diverse origins of dialogue interaction domain data, even post-cleaning, certain subsets may still degrade model performance. Verifying data quality based on final model outcomes is resource-intensive and time-consuming. We develop an efficient verification framework to overcome this. By conducting small-scale pre-training (5B tokens) and measuring real-time perplexity (PPL) changes on a dedicated validation set, we can rapidly assess a data source's contribution. This approach improves validation efficiency by over 5-fold and reduces computational costs by 95\%, enabling swift quality assessments.

\subsubsection{Ablation Studies Across Multiple Models}

We validated our continual pre-training approach on several prominent open-source models, including Yi-34B~\citep{DBLP:journals/corr/abs-2403-04652}, Qwen2.5-32B~\citep{DBLP:journals/corr/abs-2412-15115}, and our proprietary LongCat model. After training, all models demonstrated notable gains in key dialogue interaction capabilities, such as interaction proficiency and reasoning, without compromising their general performance.

As detailed in Table~\ref{tab:performance-comparison}, the improvements varied across architectures.
The evaluation metrics include \textbf{Knowledge} (accuracy on dialogue interaction multiple-choice questions), \textbf{Empathy} (quality of model-generated dialogue compared to human-annotated responses), and \textbf{Reasoning} (correctness in reasoning task outputs).
For instance, the Yi-34B-base model achieved a substantial +11.2\% overall gain, while other models showed more balanced enhancements. These results confirm the effectiveness of our method across diverse models.

\begin{table}[h]
\centering
{\footnotesize
\caption{Performance gains (\%) achieved by our continual pre-training method on dialogue interaction evaluation benchmarks across several base models.}
\label{tab:performance-comparison}
\begin{tabular}{lcccc}
\toprule
\textbf{Base Model} & \textbf{Overall Gain}~$\uparrow$ & \textbf{Knowledge Gain}~$\uparrow$ & \textbf{Empathy Gain}~$\uparrow$ & \textbf{Reasoning Gain}~$\uparrow$ \\
\hline
Yi-34B-base & +11.2 & +5.4 & +16.3 & +17.1 \\
Qwen2.5-32B-base & +3.7 & +5.6 & +5.7 & +2.0 \\
LongCat & +8.5 & +7.4 & +14.8 & +10.5 \\
\bottomrule
\end{tabular}
}
\end{table}

\vspace{-0.3cm}
\subsection{Supervised Fine-Tuning Stage}
\subsubsection{Data Paradigm Evolution}

Our objective in the SFT stage is to align the foundational model with domain-specific knowledge and intelligent interaction styles. This process reflects a shift in our data philosophy and corresponding architectural innovations. We contribute (i) a lightweight SFT paradigm emphasizing data quality over quantity, and (ii) a unified agent architecture optimized for enterprise-scale efficiency.

\paragraph{From Million-Scale to Lightweight SFT.}
Initially, we used over one million self-sampled data processed through a resource-intensive pipeline (including alignment, filtering, and cleaning) to mitigate issues like hallucinations. Although it ensured broad coverage, optimization was costly and inefficient~\citep{DBLP:conf/emnlp/WuVQH24,DBLP:journals/corr/abs-2502-04194}. We discovered that with sufficient pre-training, fewer than 10,000 high-quality samples could match the performance of our million-scale dataset. This led to a lightweight SFT paradigm, reducing training cost by 99\% and accelerating iteration from weekly to 3–4 versions per week. This agility enabled rapid data curation and architectural innovation.

\paragraph{Unified Modeling of Dialogue and Action.}
Tool use is essential in intelligent systems for acting, not just conversing~\citep{DBLP:conf/nips/SchickDDRLHZCS23,DBLP:conf/iclr/YaoZYDSN023,DBLP:conf/iclr/QinLYZYLLCTQZHT24}. Our previous decoupled design, which used separate models for dialogue and tool classification, caused latency and a ``say-do mismatch.''  Through SFT, we developed a unified model that generates both the language response and the tool command in a single JSON object. This single-step design reduces latency and hardware use while ensuring consistency between language and action, transforming the model from a conversationalist into a reliable agent.
\subsubsection{Hybrid Data-Knowledge Driven Approach}
While data-driven approaches for intelligent interaction systems achieve strong business performance through their high emotional intelligence, they suffer from heavy data-collection costs and lengthy retraining cycles whenever the knowledge bases or constraint rules are updated.
Although we adopt a lightweight SFT paradigm, it still cannot eliminate the intrinsic limitations inherent to data-driven approaches.
On the other hand, knowledge-based methods enable the development of intelligent interaction dialogue systems with enhanced flexibility in rule adaptation, improved controllability over undesirable responses, and demonstrated generalization capabilities across unseen scenarios. 
They still have some limitations. 
Specifically, in scenarios characterized by complex business processes and multiple business constraints, the conventional knowledge-driven paradigm presents substantial implementation challenges. 
These include prohibitive initial knowledge engineering costs, unsustainable maintenance requirements for comprehensive rule documentation, and persistent difficulties in ensuring the ontological completeness of the prescribed knowledge bases.

Therefore, we propose a ``data-knowledge'' dual-driven architecture which achieves a breakthrough balance between operational costs and response efficiency through organic integration of the advantages of both paradigms. 
The core innovation of this solution lies in the construction of a dynamic and adaptive ``data-knowledge'' collaboration mechanism, and the specific technical implementation can be divided into three key stages:

\textbf{Stage 1: Training Data Construction.}
The corpus comprises two complementary sources: knowledge-operational data produced by a knowledge-driven generator and a large volume of human–human dialogues that contain neither explicit knowledge nor reasoning traces.
We enforce quality through a three-tier consistency protocol: (i) self-consistency checks on the generated reasoning chains, (ii) conflict detection with constraint rules and relevant knowledge, and (iii) alignment verification between the agent response and the executable plan.
To unify the two sources, we propose an innovative fusion paradigm that unifies human-human conversational data and knowledge-operational data through prompt alignment, ensuring identical input formats. 
Concurrently, an adaptive reasoning framework enables the model to make dynamic decisions during inference: activating CoT only when retrieved knowledge is available, thus fully exploiting the complementary strengths of both data sources while efficiently utilizing computational resources.

\textbf{Stage 2: Two-Stage Optimization Approach.}
In the first stage, we perform SFT on a mixture that combines the in-domain corpus described above—operational data and human-human dialogues, which jointly account for sixty percent of the instances—with general data, making up the remaining forty percent.
We then conduct rule-based reinforcement learning on the recycled data of the deployed SFT model, which will be detailed in Section~\ref{sec:compliance}.

\textbf{Stage 3: Dual-Module Inference.}
The inference stage comprises two primary modules: knowledge retrieval and response generation.
The model assesses the available knowledge and applies it when relevant; when no knowledge is accessible, it generates responses through its internalization of abilities.
This design reduces the operational overhead of purely data-driven systems while enabling minute-level adaptation via Retrieval-Augmented Generation-based~(RAG-based) knowledge updates.

In what follows, we focus on detailing the four core components that ensure robust knowledge adherence of our intelligent interaction systems: \textbf{Prompt Engineering}, \textbf{Knowledge Production}, \textbf{Knowledge Injection}, and \textbf{Data Quality Control}. 

\paragraph{Prompt Engineering.}
To improve a model’s dialogue and knowledge compliance in complex business scenarios,
we carefully craft operational prompts that encapsulates role, instruction, constraint rules, business knowledge, solution list, system signals, and dialogue history.

\paragraph{Knowledge Production.}
During the knowledge production phase, we decompose domain expertise into minimal atomic units that are mutually independent of each other. 
For example, in the e-invoicing scenario, we refactor the original monolithic flow—where the entire procedure was encoded as a single knowledge entry—into the finest-grained facts attainable. 
This atomization improves retrieval precision and reduces downstream maintenance by enabling plug-and-play replacement of individual entries.
Each knowledge unit is required to contain three mandatory fields: (i) background context, (ii) frequent user questions, and (iii) a solution script. 
The first two delimit the scope of applicability, while the third prescribes the concrete resolution strategy.
According to cognitive complexity and content, knowledge is taxonomized into four categories: Q\&A consultation, regulatory documents, procedural workflows, and fragmentary knowledge. 
Finally, for entries that utilize external tools, we stipulate a unified tool description schema, along with a standardized output format.

\paragraph{Knowledge Injection.}
We enrich the model primarily through two strategies, batch retrieval and dialogue-based internalization, to counter insufficient knowledge recall:
\begin{itemize}[itemsep=0.5em, topsep=0.5em, label=$\blacklozenge$]
    \item \textbf{Batch Retrieval.} In batch retrieval, every knowledge unit is annotated with a scenario tag, enabling the system to fetch all entries associated with that scenario simultaneously. Although this yields a coarser granularity, it lowers retrieval difficulty and improves coverage. Because the entire scenario-specific knowledge set is returned, the model obtains a more complete view of the business context. The approach demands an extra mapping between scenarios and knowledge, yet it reduces expenditure on vector storage, computation, and maintenance.
    \item \textbf{Dialogue-Based Internalization.} For dialogue-based internalization, we note that formal knowledge is inevitably incomplete in real settings. 
    We therefore collect high-quality self-sampled data and continue to train the model on them, so that knowledge is internalized into its parameters. 
    During generation, we supply the model with a permissive prompt that encourages it to reason over the retrieved context, compare it with its parametric memory, and dynamically balance internal and external knowledge sources.
\end{itemize}

\paragraph{Data Quality Control.}
To equip the model with knowledge adherence, we iteratively built a production pipeline that generates high-quality, knowledge-grounded dialogues and injects a CoT reasoning to improve interpretability and handle complex workflows. 
First, heuristic rules combined with human annotation select human-human conversations that satisfy business constraints and fall within the model’s capacity. 
Next, an LLM-based discriminator filters retrieved knowledge for relevance. 
A second classifier checks whether the original agent response adheres to the selected knowledge, and, when necessary, rewrites the response to ensure alignment. 
For turns lacking usable knowledge, the rewrite follows the human interlocutor’s reply to minimize hallucinations. 
Finally, the entire sequence of knowledge selection, knowledge utilisation, and response construction is distilled into CoT examples that are used to fine-tune the model, endowing it with simultaneous abilities to choose appropriate knowledge and to generate compliant responses.
\subsubsection{Quality Monitoring and Data Loopback}
\begin{figure}[htbp]
\centering
\includegraphics[width=0.85\textwidth]{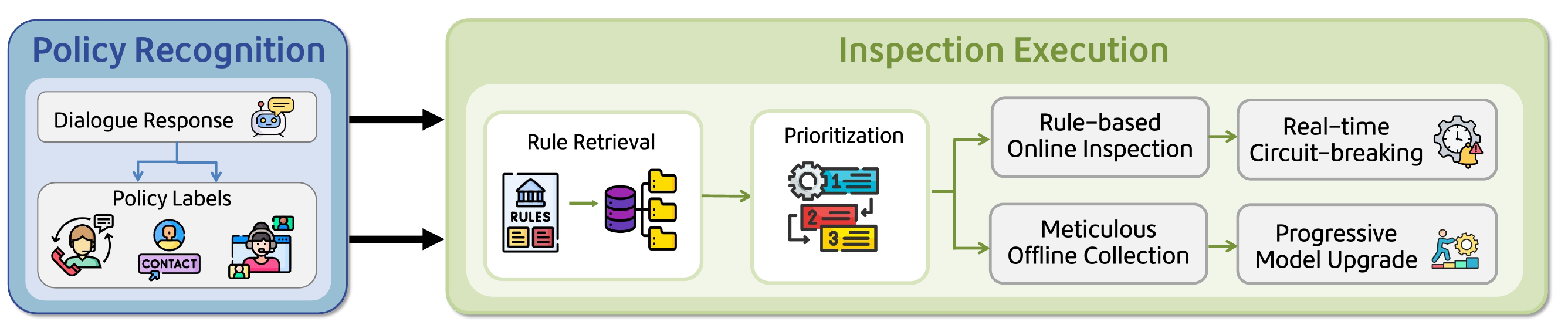}
\caption{Architecture of the data loopback framework.}
\label{fig:quality_check}
\end{figure}

The reliable performance and continuous evolution of the intelligent interaction service systems critically depend on the online quality inspection circuit breaker mechanism and the data feedback strategy. To this end, we design a quality monitoring system for data loopback, which incorporates automated quality inspection of interactive responses and curates valuable negative cases for subsequent model iterations. On the one hand, such a quality inspection mechanism should be able to ensure compliance and safety in online responses. On the other hand, the data loopback framework needs to accurately capture high-quality error cases in production environments to support the continuous optimization of the system. Our data loopback pipeline is organized into two sequential stages: (i) a Policy Recognition module and (ii) an Inspection Execution module. As shown in Figure~\ref{fig:quality_check}, the former module is responsible for routing the current dialogue scenario to different policy spaces, while the latter subsequently executes the corresponding inspections based on the specific quality-checking rules associated with each recognized policy.

\paragraph{Step 1. Policy Recognition.} Specifically, given a dialogue response and potential tool invocation from the intelligent conversational service model at a particular turn, the policy recognition module is required to identify the corresponding scenario and service policies. Notably, the policy labels (e.g., intent clarification, identity verification) are predefined and collected in advance, and a single turn of response may be associated with multiple distinct policy labels.

\paragraph{Step 2. Inspection Execution.}
The previously extracted list of action policies is then fed into the subsequent inspection-execution module. Given each action policy, a collection of rule items is then retrieved from a hierarchical multi-level rule repository for inspection, which are then prioritized accordingly (for example, an erroneous refund is considered more severe than a flawed but acceptable response). The automated quality inspection framework then evaluates these items sequentially according to their priority. Once a specific inspection item is triggered, the framework outputs the corresponding quality inspection result, including the potential error label and detailed description. Such inspection results are then collected to serve either as real-time inputs for improving model reasoning or as offline resources for model iterative upgrade.

This two-step quality inspection strategy significantly improves the accuracy of data-reflux identification. To meet the strict latency requirements of online models, we further decouple the data loopback system into two parallel mechanisms: online inspection and offline collection.
\begin{itemize}
\item {\textbf{Rule-based online inspection}}, which merely adopts simple rules that are directly executable, such as string matching or tool usage checks, to guarantee the low-latency of inspection. These results can be immediately fed back into model instructions for reflection or combined with real-time circuit-breaking strategies to maintain the reliability of online service.

\item{\textbf{Meticulous offline collection}}, which is built on multi-step, multi-agent collaborative prompt-based recognition to generate high-quality results. This process can be further enhanced with human annotation, providing strong support for model training and iteration.
\end{itemize}
Overall, this data loopback system covers 10 major inspection categories and 36 specific inspection items, addressing a wide range of known online service errors. It achieves a \textbf{93.5\%} error recall rate, with the rule-based online inspection component reaching \textbf{near-100\%} precision.
\subsubsection{Self-Refinement Training Enhancement for Intelligent Interaction Models}
With the continuous changes of business scenarios and user demands, intelligent interaction models need to possess dynamic perception and adaptability capabilities, effectively addressing the following key issues: (1) 
Training data is mainly sourced from annotated records, which may lead to limited coverage and noticeable gaps between online and offline performance~\citep{cheng2024adversarial}. (2) Existing models lack a self-iterative mechanism to adapt to new scenarios and emerging user needs. Without adapting to continuously changing real-world interactions, their performance gradually declines over time~\citep{gao2025survey}. (3) Intelligent interaction models exhibit lower solution accuracy, less anthropomorphic language expression, and limited response diversity, resulting in stiff replies, which highlights the need for targeted enhancement.

\begin{figure}[!h]
\centering
\includegraphics[width=0.85\textwidth]{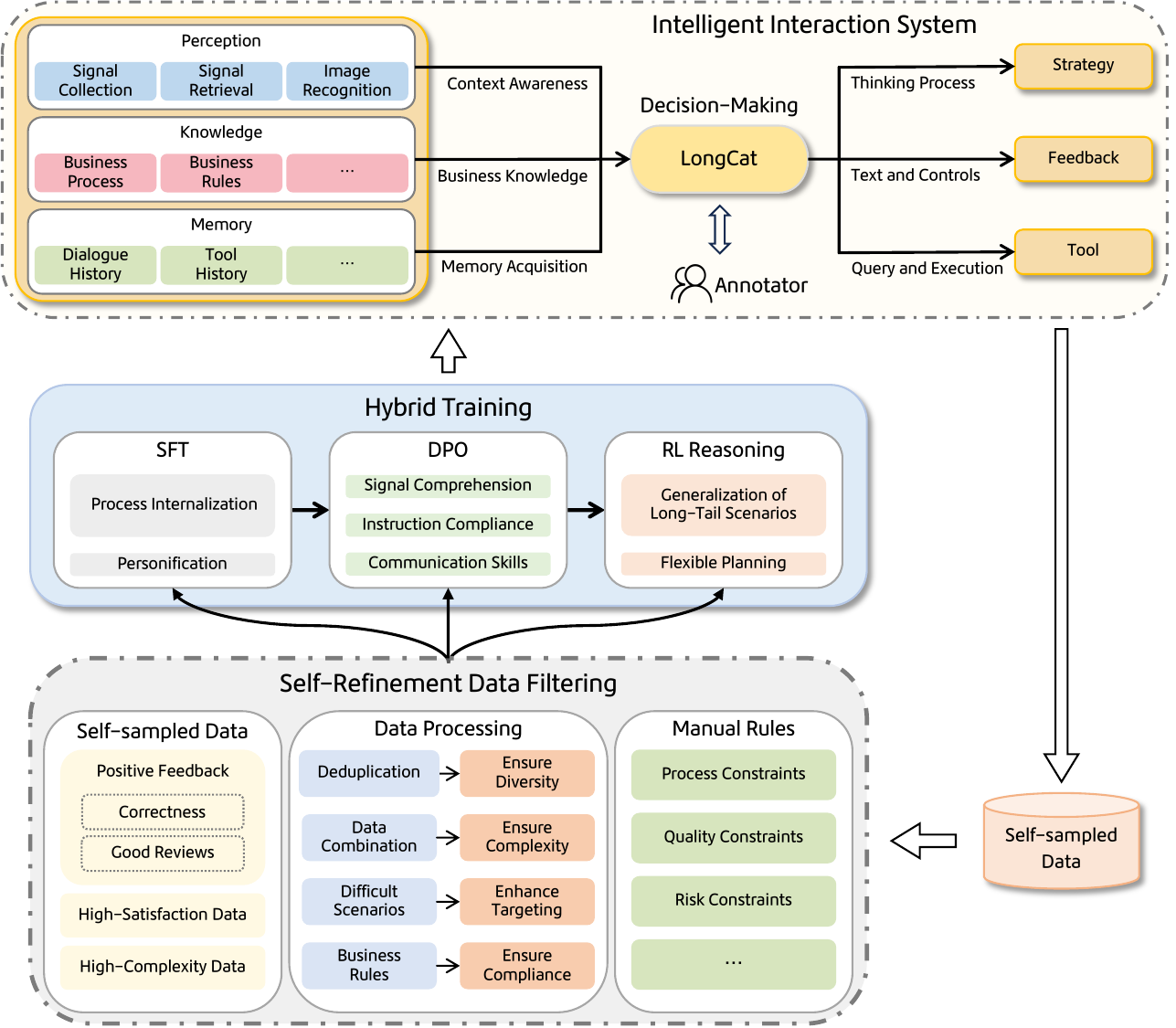}
\caption{Architecture of SRT for automated iteration using model self-sampled interaction data.}
\label{fig:srt_framework}
\end{figure}

\paragraph{SRT Framework Overview.}
To this end, we propose a Self-Refinement Training (SRT) enhancement framework, as shown in Figure~\ref{fig:srt_framework}. SRT systematically collects high-quality self-sampled cases (i.e., ``Good Cases'') to improve model response quality and achieve self-evolution. Simultaneously, it identifies and analyzes problematic cases (i.e., ``Bad Cases'') for targeted optimization, aiming to reduce manual intervention and improve overall user satisfaction. 

\paragraph{Self-Refinement Data Filtering.}
As shown in Table~\ref{tab:self_refinement_data_filtering}, SRT filters two types of self-sampled data, i.e., ``Good Cases'' and ``Bad Cases'', from three core classification indicators.
Human annotators evaluate these classification indicators for solution correctness and conversation quality, as well as simulated user models or human raters acting as users for satisfaction assessment.
A good case is defined as one where the solution is correct, the satisfaction scores from user models is high, and the conversation is of high quality. The model needs to focus on fitting this high-quality data to enable dynamic tracking and an accurate response to user demands. Bad cases primarily include two scenarios: one where the solution is correct, but receives low satisfaction scores from human evaluators or simulated user models, potentially leading to ineffective communication; and another where the solution is accurate, but the user is dissatisfied, indicating shortcomings in the models' performance in real-world applications. For these low-quality cases, the system automatically identifies and classifies them, using targeted rewriting for optimization.

\begin{table}[!h]
\centering
{\footnotesize
\caption{Self-refinement data filtering criteria for good cases and bad cases.}
\label{tab:self_refinement_data_filtering}
\begin{tabular}{@{}ccccc@{}}
\toprule
\textbf{Solution Correctness} & \textbf{User Satisfaction} & \textbf{Conversation Quality} & \textbf{Category} & \textbf{Usage} \\
\midrule
\ding{52} & \ding{52} & \ding{52} & Good Cases & SFT \\
\ding{52} & \ding{52} & \ding{56} & Bad Cases & DPO/RL \\
\ding{52} & \ding{56} & \raisebox{0.5ex}{\rule{0.15cm}{1pt}} & Bad Cases & DPO/RL \\
\ding{56} & \raisebox{0.5ex}{\rule{0.15cm}{1pt}} & \raisebox{0.5ex}{\rule{0.15cm}{1pt}} & - & Unused \\
\bottomrule
\end{tabular}
}
\end{table}

\paragraph{Self-Evolution Based on Good Cases.}

SRT automatically retrieves and preprocesses self-sampled data, applying predefined rules to ensure format consistency. To maintain data diversity, SRT utilizes stratified sampling to balance the representation across different solution types. In the data refinement stage, the pre-processed dialogues undergo multi-dimensional service quality scoring, retaining only high-quality dialogues that meet or exceed human-level service standards. Finally, knowledge-driven solution correctness verification ensures that all selected samples are valid, comprehensively improving the quality and richness of the model training data, thus enabling the self-evolution of intelligent interaction models. Table~\ref{tab:srt_model_comparison_1} shows the performance comparison of different models, which focus on two user satisfaction metrics, i.e., USM 1 and USM 2.

\begin{table}[!h]
\centering
{\footnotesize
\begin{tabular}{lcccccc}
\toprule
\textbf{Model} & \textbf{USM 1 $\downarrow$} & \textbf{USM 2 $\uparrow$} \\
\midrule
Base                         & 52.91                       & 27.21  \\
SFT                & $38.56^{-14.35}$            & $40.61^{+13.40}$  \\
SRT w/ Good Cases  & $\mathbf{28.18^{-24.73}}$   & $\mathbf{48.15^{+20.94}}$  \\
\bottomrule
\end{tabular}
}
\caption{Performance comparison on the LongCat model based on USM 1 (\%) and USM 2 (\%). For USM 1, a lower score is better; for USM 2, a higher score is better.}
\label{tab:srt_model_comparison_1}
\end{table}

\paragraph{Targeted-Learning Based on Bad Cases.}
To further improve the utilization efficiency of self-refinement data, SRT utilizes bad cases for targeted learning. Specifically, SRT analyzes sampled data to identify problematic dialogue samples (such as repetitive conversation or awkward expressions), and then uses a rewriting model (open-source LLMs trained via SFT) to deduplicate and optimize these samples. Furthermore, the original and rewritten samples can be used to form preference pairs for subsequent DPO or RL training, thus continuously improving the service quality of intelligent interaction models. In Table~\ref{tab:srt_model_comparison_2}, we demonstrate the advantages of this optimization in terms of two user satisfaction metrics (USM 1 and USM 2) and repetition rate (RR).

\begin{table}[!h]
\centering
{\footnotesize
\begin{tabular}{lccccccc}
\toprule
\textbf{Model} & \textbf{USM 1 $\downarrow$} & \textbf{USM 2 $\uparrow$} & \textbf{RR $\downarrow$} \\
\midrule
Base                                      & 52.91                      & 27.21                          &57.43 \\
SRT w/ Good Cases               & $28.18^{-24.73}$           & $48.15^{+20.94}$               &$45.12^{-12.31}$ \\
SRT w/ Good Cases \& Bad Cases  & $\mathbf{25.38^{-27.53}}$  & $\mathbf{52.72^{+25.51}}$      &$\mathbf{20.27^{-37.16}}$ \\
\bottomrule
\end{tabular}
}
\caption{Performance comparison on the LongCat model based on USM 1 (\%), USM 2 (\%), and RR (\%). For USM 1 and RR, a lower score is better; for USM 2, a higher score is better.}
\label{tab:srt_model_comparison_2}
\end{table}

\subsection{Preference Learning Stage}
Preference learning~\citep{christiano2017deep} is a crucial method that effectively aligns language models with more human-friendly outputs, and DPO~\citep{rafailov2023direct}~ serves as a key technique to achieve this goal. It uses comparative human feedback data—pairs of model responses ranked as ``preferred'' or ``dispreferred'' by annotators—to fine-tune models directly, removing the need for intermediate reward model training~\citep{schulman2017proximal}, thus simplifying the alignment process and boosting efficiency. The formula for the DPO loss is as follows:

\[
\mathcal{L}_{\text{DPO}}(\theta) = -\mathbb{E}_{(x, y_+, y_-) \sim \mathcal{D}} \left[ \log \left( \sigma \left( \frac{1}{\beta} \left( \log P_\theta(y_+ \mid x) - \log P_\theta(y_- \mid x) \right) \right) \right) \right].
\]

For LLMs, the Preference Dataset ($\mathcal{D}$) consists of triples ($x, y_+, y_-$), where $x$ denotes input prompts, $y_+$ stands for the higher-quality response, and $y_-$ represents the inferior response; Model Probability ($P_\theta(y \mid x)$) is the conditional probability that an LLM (parameterized by $\theta$) generates response $y$ given input $x$, reflecting the model's inherent tendency to produce $y$ for $x$; Temperature Hyperparameter ($\beta$) regulates the LLM's sensitivity to quality differences between $y_+$ and $y_-$, and modulates the prioritization of their probability gap during optimization; the Sigmoid Function ($\sigma(\cdot)$), mathematically expressed as $\sigma(z) = \frac{1}{1 + e^{-z}}$, is a logistic function mapping real $z$ to [0,1], and in this context, it takes the normalized (scaled by $1/\beta$) log-probability difference of $y_+$ and $y_-$ as input, converting it to a probability-like value to guide maximizing the likelihood of preferred responses.

In intelligent interaction systems, LLMs interact frequently with users on diverse inquiries. The quality of their outputs—clarity, professionalism—directly impacts user satisfaction; poor responses can quickly harm the user experience. Hence, integrating human preference data via methods like DPO is critical: it captures the subtle standards of good intelligent interaction systems, thereby enabling models to generate more acceptable responses that better meet real-world needs.

We developed a standardized DPO workflow, including annotation guidelines, scenario-based data selection, and optimized fine-tuning parameters. This has continuously driven significant progress in improving dialogue interactions. In some challenging scenarios, the performance is far better than before: higher problem-solving efficiency, better satisfaction scores.

\subsubsection{Domain-Specific issues}
In practical application, our model has encountered numerous issues. These problems not only prevent users' concerns from being resolved promptly and correctly, but also result in poor user experience, and may even mislead users and lead to unforeseen consequences. To systematically analyze and address these issues, we have conducted an in-depth analysis and summary, identified their common causes and characteristics, and plan to resolve them through the DPO method. The following is an explanation of these problems along with corresponding examples.

\begin{itemize}
\item \textbf{Unclarified User Demands: }
The model fails to clarify user demands. 

\item \textbf{Redundant Inquiries: }
The model repeatedly asks users for information that has already been provided by the users themselves or acquired by the model from external sources.

\item \textbf{Hallucinations: }
The model responds to users with false or unsubstantiated information.

\item \textbf{Script Repetition: }
The model continuously repeats previously used script content and is unable to resolve user problems or soothe their emotions effectively.

\item \textbf{Incorrect Solutions: }
The model provides incorrect methods in response to user demands.

\item \textbf{Failure to Pay Due Compensation: }
In scenarios where proactive compensation is applicable, the model fails to propose compensation, resulting in a poor user experience.

\item \textbf{Compliance with Compensation Signals: }
When compensation signals are clear, the model fails to comply with these signals, leading to a subpar user experience.

\item \textbf{Failure to Advance the Process: }
The model cannot effectively advance the handling process. 

\end{itemize}

\subsubsection{Operational DPO}

\begin{figure}[htbp!]
    \centering
    \includegraphics[width=0.8\textwidth]{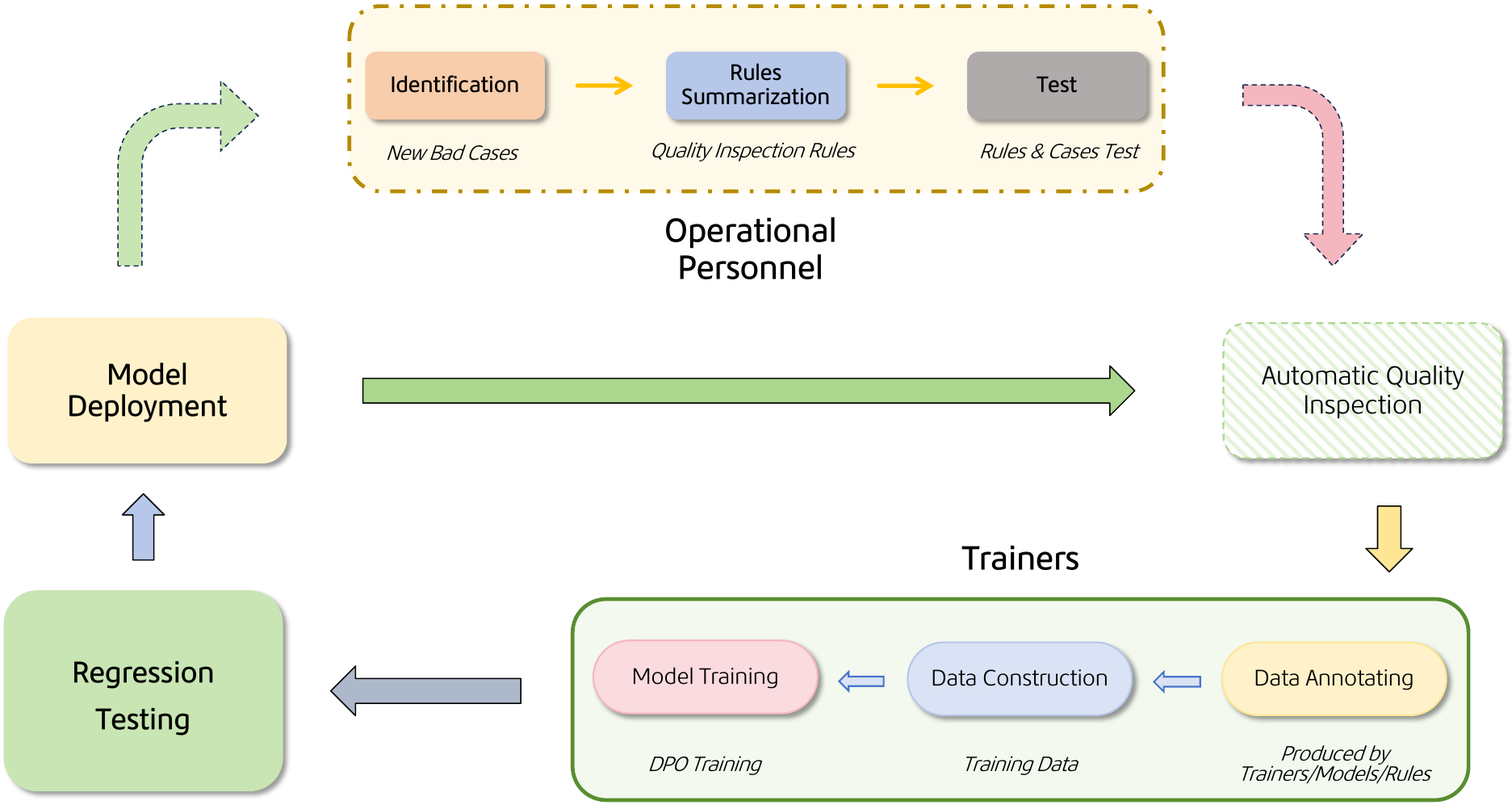} 
    \caption{Schematic Diagram of the Operable DPO Framework. Operational Personnel, trainers, and automated tools each participate in various coordination aspects throughout the entire process and collaborate with one another, thereby establishing an efficient and sustainable iterative process.}
    \label{fig:operational_dpo}
\end{figure}

To effectively mitigate the aforementioned issues (e.g., unclarified user demands, redundant inquiries, hallucinations) and fundamentally improve user satisfaction in interactive scenarios, we developed a DPO-based framework tailored for further fine-tuning LLMs. This framework aims to align the model’s responses with the quality, empathy, and problem-solving capabilities of human interaction representatives, with ultimate goals of boosting user demand resolution rates, reducing human transfer frequency, and ensuring consistent, effective model performance in real-world applications. Subsequent sections will elaborate on the framework’s key components.

Figure~\ref{fig:operational_dpo} illustrates the operational DPO framework. The process starts with continuous identification of bad cases. Next, quality inspection rules for these bad cases are analyzed and summarized. A series of tests then verify rule compliance and avoid anomalies: edge cases are fully considered for regex extraction, while prompts are continuously optimized for LLM-based detection to ensure accurate identification. Simultaneously, real-time bad case mining is conducted via automatic quality inspection of self-sampled data. Subsequently, data annotation is performed using a combination of annotators, models, and rule-based systems;after screening and further processing, training data construction is completed. Once training data is ready, the DPO model training is initiated to enable the model to learn the latest rules. Finally, regression testing (including automatic and manual evaluation) is implemented to ensure the updated model performs normally on historical data without ``catastrophic forgetting''; the latest model version is then deployed, starting the next iteration.

\begin{table}[h]
\centering
{\footnotesize
\begin{tabular}{@{}lcc@{}}
\toprule
\textbf{Issue name} & \textbf{Base $\uparrow$} & \textbf{DPO $\uparrow$} \\ 
\midrule
Unclarified User Demands & 34.52 & 40.70\textsuperscript{+6.18} \\ 
Redundant Inquiries & 73.68 & 78.94\textsuperscript{+5.26}  \\ 
Hallucinations & 85.00 & 97.50\textsuperscript{+12.50} \\ 
Script Repetition & 89.54 & 98.25\textsuperscript{+8.71} \\ 
Incorrect Solutions & 65.51 & 100.00\textsuperscript{+34.49} \\ 
Failure to Pay Due Compensation & 27.77 & 29.62\textsuperscript{+1.85} \\ 
Compliance with Compensation Signals & 76.36 & 81.50\textsuperscript{+5.14} \\ 
Failure to Advance the Process & 83.41 & 100.00\textsuperscript{+16.59} \\ 
\bottomrule
\end{tabular}
}
\caption{The repair rate of bad cases in domain-specific issues. ``Base'' refers to the control group model, i.e., the version before the latest DPO fine-tuning; ``DPO'' denotes the latest model iterated using our operational DPO method. All values in this table are expressed as percentages (\%).}
\label{tab:badcase_repair_rate}
\end{table}

The entire process requires the coordinated effort of operations personnel, trainers, and various automated tools, thereby establishing a robust iterative loop. This framework facilitates the gradual learning and growth of the intelligent interaction systems during its actual deployment and usage, enabling continuous performance improvement. Ultimately, it aims to achieve a significantly high level of user satisfaction and problem resolution rate.

Table \ref{tab:badcase_repair_rate} presents the repair rate of bad cases addressed by our method in the aforementioned special problem scenarios, which refers to the proportion of samples where these issues were successfully resolved. It can be observed that with the continuous iteration of the operational DPO framework, the repair rate for various types of problems has shown a consistent upward trend. In certain scenarios—such as ``Hallucinations''—the model's repair rate has even reached 97.5\%. In the ``Incorrect Solutions'' scenario, the improved method has achieved an increase of 34.49\% in the repair rate, which convincingly demonstrates the effectiveness of this approach.

\subsection{Reasoning Enhancement Stage}
\subsubsection{Reinforced Reasoning Enhancement for General Dialogue and Knowledge Following Ability}
\label{sec:compliance}
\label{sec:label_2_4_1_Enhancing}
In this section, we present our approaches to enhancing both the general conversational capability and knowledge following of the model. Our discussion proceeds in two progressive steps. First, we improve the model’s ability to follow knowledge under multiple constraints, thereby further enhancing its general conversational performance. Second, we conduct practical reinforcement of reasoning through hybrid data- and knowledge-driven approaches in the intelligent interaction domain, aiming to optimize the model’s real-world application.

\paragraph{Enhancing General Conversational Ability through Multi-Constraint Knowledge Following.}
Many high-value business scenario instructions often exhibit complex characteristics such as multi-business constraints, compound tasks, and cumbersome business processes~\citep{qu2025survey}. Examples include smart outbound call robots and smart intelligent interaction robots~\citep{wulf2024exploring,akhtar2024developing}. However, many open-source/self-developed large models reveal issues such as non-compliance with format requirements and business constraints, hindering ideal effectiveness in business scenarios~\citep{lou2024large,qin2024infobench,so2025large}. To improve a model's dialog and knowledge compliance in complex business scenarios, we aim for targeted optimization of self-developed models' shortcomings in following complex instructions for wider deployment and benchmark alignment with industry standards.
\begin{figure}[t]
    \centering
    \includegraphics[width=0.8\textwidth]{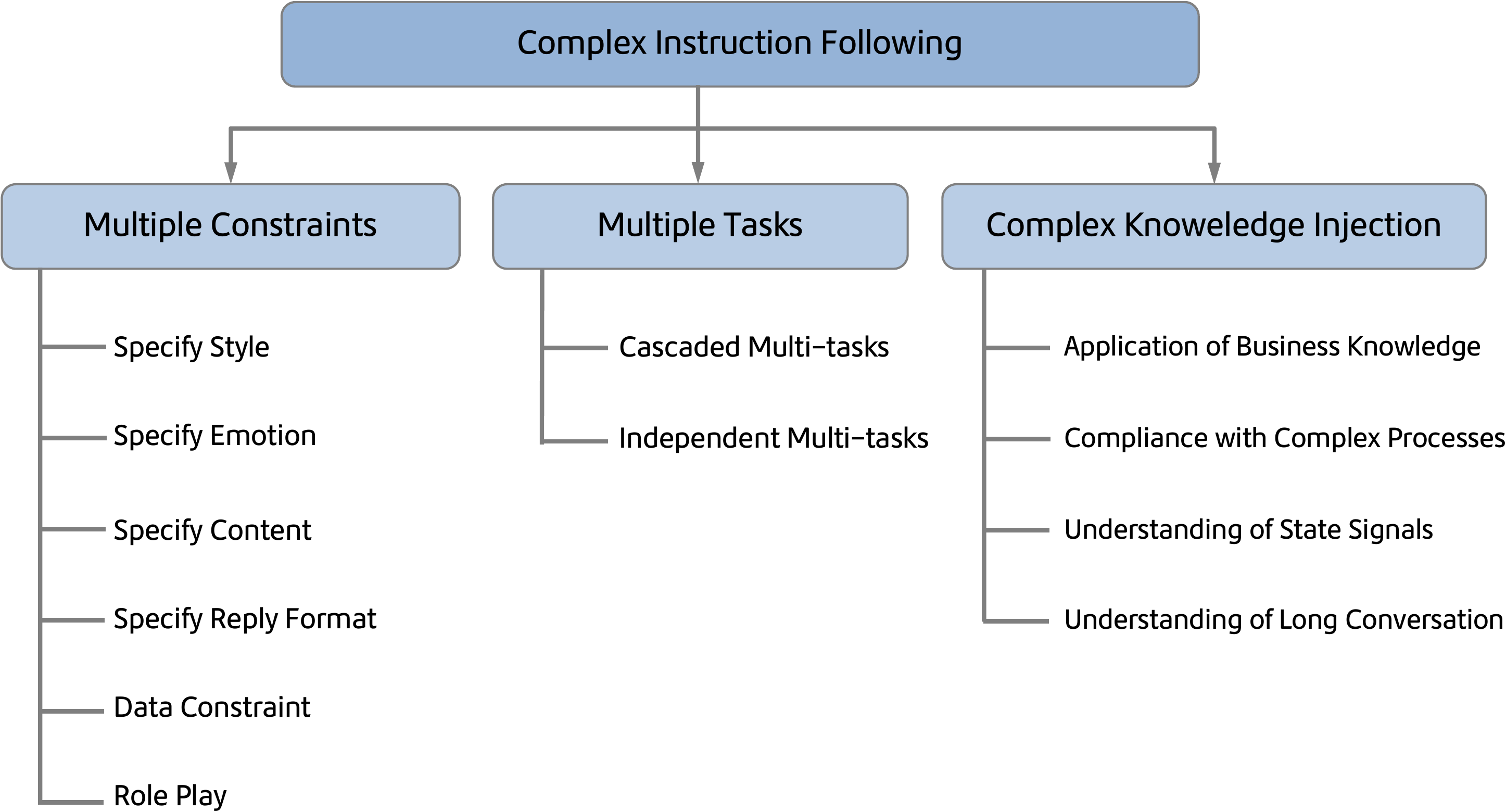}
    \caption{The Category of Complex Instruction.}
    \label{fig:if_category}
\end{figure}

High-quality complex instruction data in Chinese is limited and exhibits significant differences from instructions commonly found in real-world business scenarios, which results in the direct use of open-source datasets being ineffective for improving our models. Drawing upon academic categorizations of complex instructions~\citep{zhang2024cfbench,wen2024benchmarking}, we incorporate the unique characteristics of Meituan's business scenarios to classify complex instructions into three major categories: multi-constraint, multi-task, and complex knowledge integration. Furthermore, we refine these categories into twelve distinct subcategories, enabling more targeted data mining and synthesis to better address the diverse needs of our applications, shown in Figure~\ref{fig:if_category}.

\textbf{Data Mining and Synthesis:} Utilizing complex instruction categories and definitions, we apply LRM for automated labeling and mining from large business instruction datasets. For missing types, data is synthesized via autonomous creation, complexity evolution, and integration, guided by personas, branch count, constraints, and task types to address template challenges. Seed data is scored for correctness, completeness, and conciseness, discarding low-quality or illogical instructions. An objective tool for evaluating difficulty is needed to filter high-value data for model improvement.

\textbf{Difficulty Assessment and Optimization:} We assess instruction difficulty by sampling responses and voting across models with varying capabilities, removing overly simple data with similar distributions to preserve overall difficulty. Open-source benchmark models like DeepSeek-R1~\citep{guo2025deepseek} are used for generating responses and thought chains. For verified data, we compare R1 predictions with standards, selecting consistent thought chains; for unverified data, multiple R1 responses are sampled, with models selecting the best. This process yields high-quality complex instruction data, boosting model comprehension and knowledge alignment.

\textbf{Reinforcement Phase:} Most complex instructions lack standard answers, making rigid validation unfeasible. For verifiable data, rule matching calculates rewards; for non-verifiable data, DeepSeek’s Self-Principled Critique Tuning~(SPCT)~\citep{liu2025inference} enables principle-based reinforcement. SPCT develops evaluation principles, generating reward scores automatically. Unlike GRM~\citep{mahan2024generative}, principle generation and judgment are decoupled, using DeepSeek-R1 pre-training to enhance principle quality and lower RL training costs significantly. Qwen2.5-32B~\citep{team2024qwen2} is used as GRM for principle-based scoring during training.

On Qwen2.5-32B, inference tuning with reinforcement achieves industry-leading levels on CFBench and our complex business instruction compliance benchmark, detailed in Table~\ref{tab:benchmark_performance_table6}. 

\begin{table}[t]
\centering
{\footnotesize
\begin{tabular}{@{}l>{\centering\arraybackslash}m{2.2cm}cc@{}}
\toprule
\textbf{Model} & \textbf{Average Pass@1} & \multicolumn{2}{c}{\textbf{Benchmark}} \\ 
\cmidrule(lr){3-4}
 &  & \textbf{CFBench} & \textbf{Complex\_Bench\_ModelEval} \\ 
\midrule
Deepseek-R1 & 73.56 & 76.33 & 70.79 \\ 
QwQ-32B & 71.52 & 74.33 & 68.71 \\ 
Qwen3-235B-A22B & 70.94 & 75.00 & 66.87 \\ 
Qwen3-32B & 71.68 & 75.33 & 68.03 \\ 
GPT4o-1120 & 71.46 & 74.30 & 68.62 \\ 
Qwen2.5-32B & 64.52 & 64.33 & 64.70 \\ 
Qwen2.5-32B w/SFT & 66.96 & 69.00 & 64.92 \\ 
\textbf{Qwen2.5-32B w/SFT+RL} & 72.88 & 74.89 & 70.86 \\
\bottomrule
\end{tabular}
}
\caption{Benchmark performance (Pass@1) across various models.}
\label{tab:benchmark_performance_table6}
\end{table}

\paragraph{Hybrid Data-Knowledge Driven Approach for Intelligent Interaction Models.}
Intelligent interaction systems require both cognitive and emotional intelligence~\citep{maroju2025understanding,wang2024research,singh2024human}, demanding the ability to understand complex business logic, process real-time signals in multi-turn dialogues, and generate accurate, fluent responses. Key challenges include: (a) base models struggle to balance complex business comprehension with natural interaction, and (b) business process complexity limits optimization via short-term measures, complicating model adaptation. Traditional data-driven methods are constrained by the need for extensive high-quality training data and inefficient, costly error correction, often necessitating frequent retraining.

We introduce a hybrid method integrating data-driven and knowledge-based techniques to improve business instruction understanding, reduce rote memorization, and enable rapid, effective error correction through recall of relevant business knowledge. This strategy enhances model efficiency and accuracy while reducing retraining frequency.

\begin{figure}[htbp]
    \centering
    \includegraphics[width=0.8\textwidth]{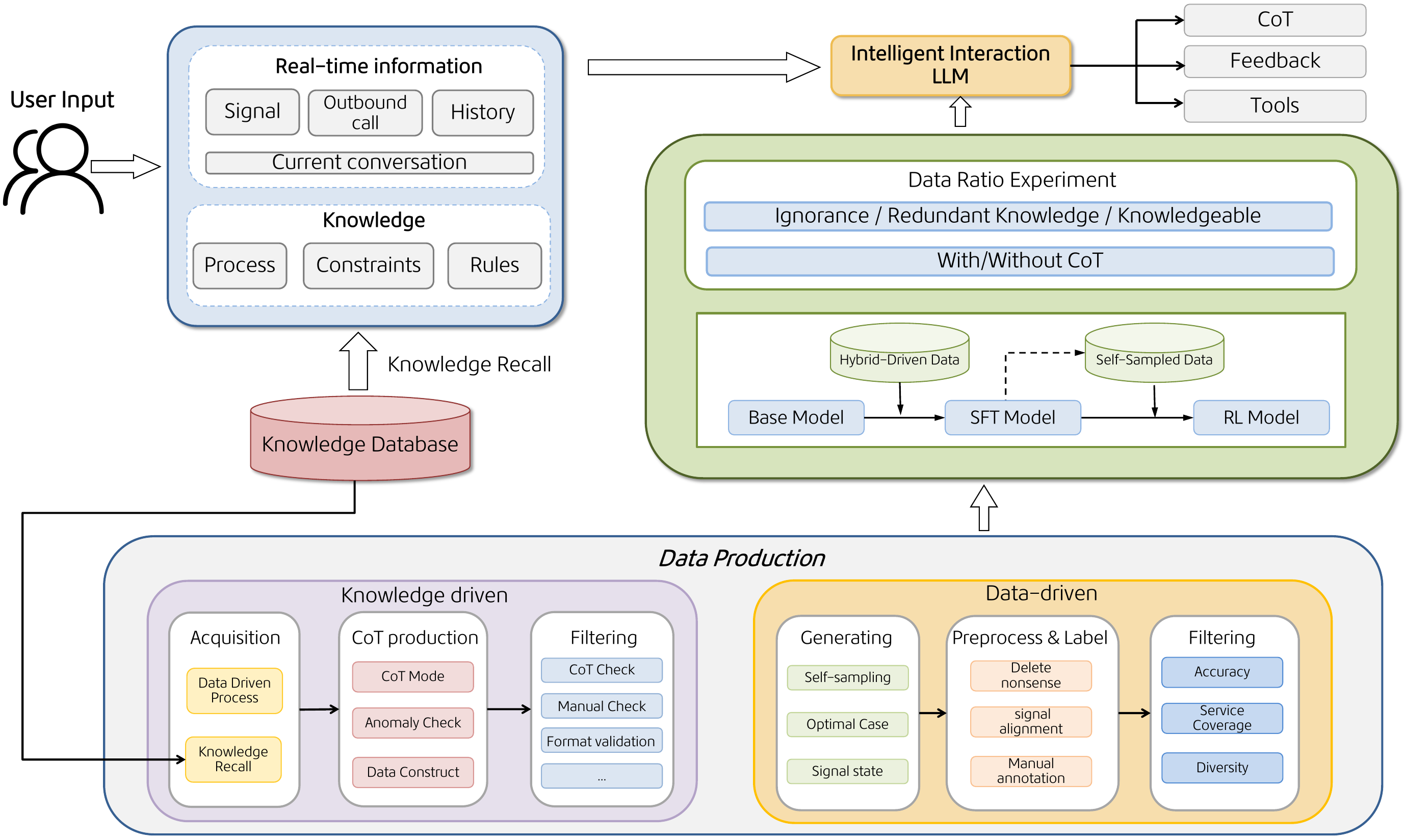} 
    \caption{We explore both data and technical aspects to develop a ``knowledge-data hybrid-driven'' process. This combines structured business knowledge with large-scale datasets. We focus on ``data ratio optimization'' and ``reinforcement learning strategies'' to improve the model's adherence to knowledge-based rules, enabling better handling of complex business scenarios.}
    \label{fig:hybrid_training_process}
\end{figure}

\textbf{Model Design and Framework Overview:}
The core challenge in building a ``data + knowledge operational'' hybrid model is enabling independent assessment of knowledge availability. The model should retain data-driven strengths, mastering standard business processes and ensuring smooth dialogue, while overcoming their limitations by flexibly applying business knowledge to adapt to changing requirements. The hybrid architecture allows the model to assess and utilize available knowledge; if none is found, it reverts to data-driven response generation, detailed in Figure~\ref{fig:hybrid_training_process}.

\textbf{CoT Data Generation:}
In the ``knowledge-data''-driven CoT~\citep{mann2020language,wang2022self} production process, we must meet two main requirements. 
First, the reasoning path must be clear and designed according to actual business logic. 
Second, the reasoning content must be accurate, with all information in the CoT being verifiable in the input data to avoid fabrications. 
Given the high cost and homogeneity of manual annotation, which makes it difficult to ensure CoT diversity, we use an ``automated generation + two-step validation'' approach. First, the model generates CoTs in bulk. 
Then, it performs both forward and reverse validation of the referenced knowledge. The filtered CoTs are reviewed by annotators, and any errors are removed.

\textbf{Rule-Based Reinforcement Learning:}
We begin by performing a cold start on the model using SFT~\citep{wei2025advancing}. 
The training set includes ``data-driven'' and ``knowledge-driven'' data with CoT reasoning paths, guiding the model's basic training. 
This stage helps the model learn basic intelligent interaction dialogue and the reasoning paths behind business knowledge, transitioning from general to business-specific capabilities. 
Afterward, based on the SFT model, we reintroduce ``knowledge-driven'' data with CoT reasoning paths and incorporate reinforcement learning.

A reward function $R_{total}$ is designed to evaluate the model's output against high-quality intelligent interaction processes, driving the model to adjust its strategies and optimize outputs. This continuous refinement improves output quality and logical consistency:
\[
R_{\text{total}} = R_{\text{sol}} + R_{\text{kn}} + R_{\text{dlg}} + R_{\text{cot}},
\]
where each component is defined as follows:
\begin{itemize}[itemsep=0.5ex, parsep=0pt]
    \item \textbf{Solution Correctness ($R_{\text{sol}}$, 0.1/1)}: Equals 1 if the model’s proposed solution matches the standard answer; otherwise 0.1.
    
    \item \textbf{Knowledge Reference Correctness ($R_{\text{kn}}$, 0.1/1)}: Equals 1 if the model uses the correct knowledge reference ID; otherwise 0.1.
    
    \item \textbf{Dialogue Appropriateness ($R_{\text{dlg}}$, 0.1–1)}: This is determined by two sub-components: repetition penalty and answer similarity.
    
    \item \textbf{Repetition Penalty ($R_{\text{rep}}$)}: If the Jaccard similarity between the generated response and historical dialogue is greater than $0.8$, or the longest common substring ratio is greater than $0.5$ with length greater than $18$, then $R_{\text{rep}} = 0.1$; otherwise $R_{\text{rep}} = 1$.
    
    \item \textbf{Answer Similarity ($R_{\text{sim}}$)}: Using the BGE model, the similarity score $s$ is computed as follows:
    $$
    R_{\text{sim}} =
    \begin{cases}
        0.1, & s \leq 0.65, \\
        1, & s \geq 0.9, \\
        f(s), & \text{otherwise},
    \end{cases}
    $$
    where $f(s)$ is an exponential scaling function.
    
    \item \textbf{CoT Length Penalty ($R_{\text{cot}}$, 0.1/1)}:
    $$
    R_{\text{cot}} =
    \begin{cases}
        0.1, & \text{if length} > 275, \\
        \max\left(1 - \left(\frac{0.6 - \frac{L_{\text{gen}}}{L_{\text{ans}}}}{0.6}\right) \times 2, 0.1\right), & \text{if } L_{\text{gen}} < 0.6 \cdot L_{\text{ans}}, \\
        1, & \text{otherwise},
    \end{cases}
    $$
    where $L_{\text{gen}}$ is the length of the generated rollout, and $L_{\text{ans}}$ is the length of the ground truth.
\end{itemize}

To evaluate the effectiveness of the proposed method, we conducted experiments in one of the rider scenarios. Table~\ref{tab:performance_comparison_table7} presents the accuracy performance of different training strategies. The results show that reinforcement learning (RL and RL-CoT) achieves the highest accuracy, with an improvement of 1.05 percentage points over the baseline, demonstrating the superiority of our approach in enhancing model performance for this business scenario.

\begin{table}[htbp]
    \centering
    {\footnotesize
    \begin{tabular}{lccccc}
        \toprule
        \textbf{Methods} & Baseline & SFT & SFT-CoT & RL-CoT & RL \\
        \midrule
        Accuracy & $70.83$ & $71.35^{+0.52}$ & $71.35^{+0.52}$ & $71.88^{+1.05}$ & $71.88^{+1.05}$ \\
        \bottomrule
    \end{tabular}
    }
    \caption{Performance comparison of different training methods in one of the rider scenarios. The upper part shows the relative accuracy improvement over the baseline, while the lower part reports the absolute accuracy values(\%).}
    \label{tab:performance_comparison_table7}
\end{table}

\subsubsection{Reinforcement Finetuning for Enhanced Humanized Dialogue and Emotional Care}
Section~\ref{sec:label_2_4_1_Enhancing}~has discussed methods for enhancing the model’s general conversational ability and knowledge compliance. However, in intelligent interaction systems, a qualified intelligent agent must demonstrate not only strong conversational skills but also a high degree of humanization and emotional care. Therefore, this section focuses on creating a more empathetic, intelligent interaction experience by improving the model’s human-like and caring attributes.

Our enhancement approach consists of two aspects:  
(a) First, we leverage the GRM to assess dialogue quality across multiple dimensions, thereby guiding the policy model to optimize for greater humanization.  
(b) Second, we employ a dialogue rewriting model to further improve the quality and emotional expressiveness of intelligent interactions.

\paragraph{GRM-based Humanization of Dialogue in Open-Ended AI Intelligent Interaction.}
Currently, rule-based reinforcement learning methods are effective for tasks with clear reward signals.
However, in open-ended dialogue scenarios~\citep{xue2025mmrc,liang2024mathchat}, the difficulty in evaluating model outputs leads to inaccurate reward signals, which can mislead the optimization direction. 
Traditional SFT models often suffer from repetitive and rigid dialogue, making it difficult to generate creative and fluent responses. 
Given the vast output space of open-ended dialogue tasks, it is impossible to assign precise rewards to every response using simple rules, resulting in the model being susceptible to noise in the reward signal, potentially optimizing in the wrong direction.

To fully leverage the potential of reinforcement learning, it is essential to refine the reward design so that it closely aligns with human evaluation standards, thereby improving dialogue flexibility and practical application performance.
Consequently, we introduce a more refined Reward Model, which significantly enhances consistency with human annotations in response evaluation tasks. By integrating this with RL training, the overall quality of model-generated dialogue is further improved.

\begin{figure}[htbp]
    \centering
    \includegraphics[width=0.8\textwidth]{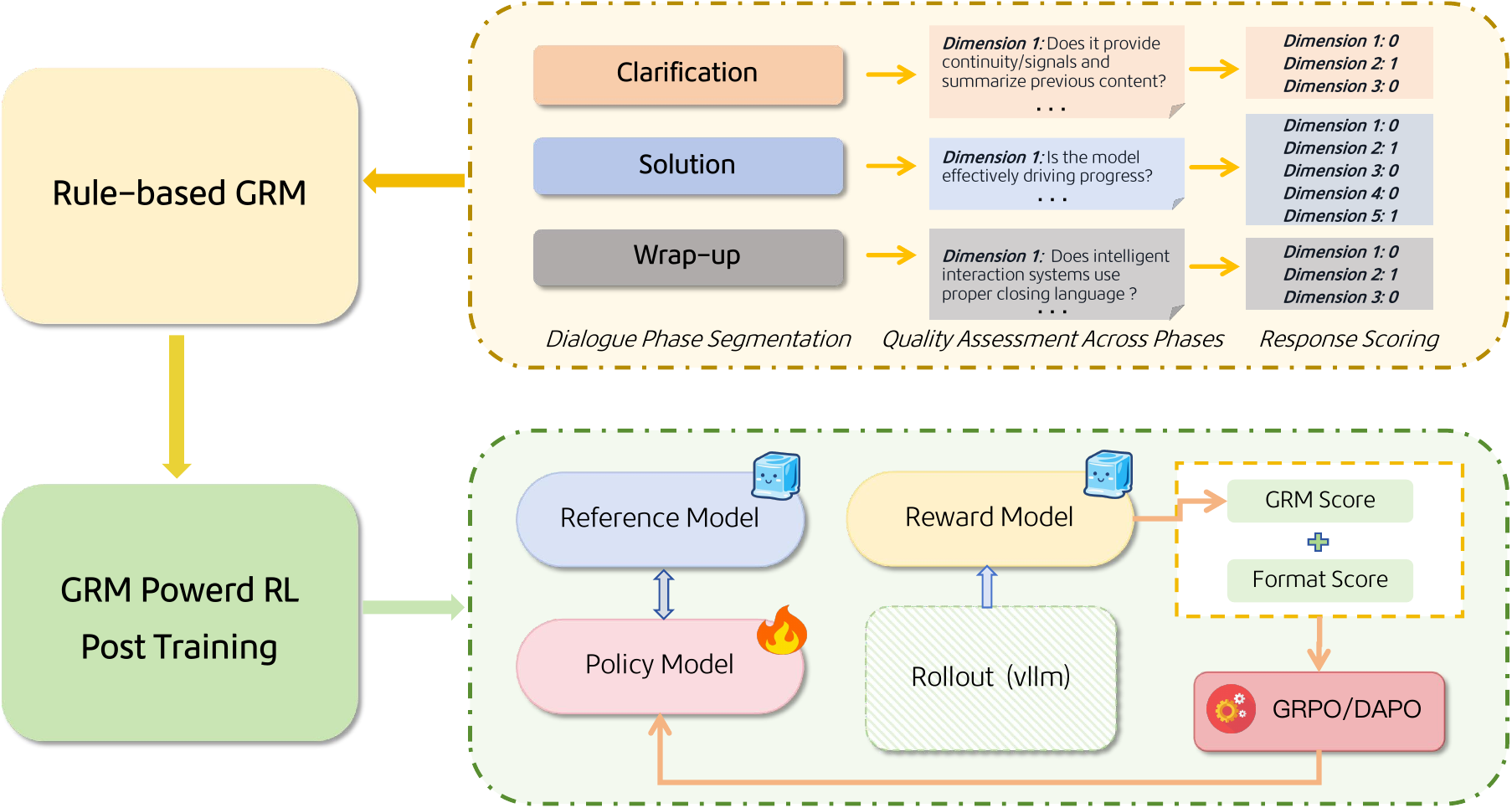}
    \caption{Architecture of GRM-Powered RL System with Multi-Phase Dialogue Assessment.}
    \label{fig:grm_architecture}
\end{figure}

\textbf{Rule-based Fine-tuning Generation Reward Model:}
Since BT training methods~\citep{mckinney2023fragility,ankner2024critique} require substantial amounts of data, we opted for using the GRM~\citep{liu2025inference,yu2024self,chen2025rm} to construct the reward model, achieving more efficient and precise optimization. 
The GRM-based reinforcement fine-tuning reward model utilizes preference data produced under high-quality dialogue standards for RL training. 
Regarding the data processing, we extracted historical dialogues from self-sampled data. Due to the specific nature of these dialogues, we divided them into three stages, as shown in Figure \ref{fig:grm_architecture}. 
Each stage has its own unique dimensions and evaluation criteria, providing a binary score of 0 or 1. 
Finally, these scores are normalized to a floating-point number between 0 and 1.
This model has successfully achieved over 85\% consistency with human annotations, even when working with a limited dataset through RL.

Upon examining the human-annotated data, we found that most of the examples were positive, and the model, after training, tended to assign positive scores. To mitigate this bias, we proposed a reward function to adjust the output scores and correct for the overemphasis on positive responses.
For each sub-dimension, the reward \( R \) is designed as follows:

\[
R =
\begin{cases} 
r_{00}, & \text{if } g_t = 0, \, r_s = 0, \\
-r_{01}, & \text{if } g_t = 0, \, r_s = 1, \\
+r_{11}, & \text{if } g_t = 1, \, r_s = 1, \\
-r_{10}, & \text{if } g_t = 1, \, r_s = 0,
\end{cases}
\]
where \( g_t \) represents the true label and \( r_s \) represents the model output. To ensure that reinforcement learning better adapts to the specific requirements of each task, the values of \( r_{00} \), \( r_{01} \), \( r_{11} \), and \( r_{10} \) should be set differently according to the task distribution. By assigning distinct weights to each reward term based on the prevalence and importance of each outcome in a given task, we can emphasize the correction of certain types of errors or successes. This flexible weighting mechanism enables the reward function to guide the model toward optimal behavior for diverse task scenarios, improving both accuracy and robustness in reinforcement learning training.

\textbf{Trained GRM and RL Fine-tuning Policy Model:}
After training the GRM, we integrate it with RL to perform policy model reinforcement training.  
This policy model is designed to further refine dialogue strategies through reinforcement learning.  
Our objectives are to reduce handovers to human agents and increase the overall resolution rate, making multi-turn dialogues more human-like and ensuring smoother, more natural context continuity.  
As shown in Table~\ref{tab:model_comparison_grm_table8}, the trained GRM with RL surpasses both the baseline BGE and the untrained GRM in key metrics, including full speech score rate, usability rate, and solution accuracy.

\begin{table}[htbp]
    \centering
    {\footnotesize
    \begin{tabular}{lcccccc}
        \toprule
        \textbf{Model} & \textbf{FSR} & \textbf{UR} & \textbf{BI} & \textbf{Acc} & \textbf{SR} \\
        \midrule
        BGE & 62.86 & 65 & 0 & 79.09 & 78.57 \\
        \midrule
        BGE + Offline Training & 65.71 & 67.86 & 0 & 78.3 & 75.71 \\
        \midrule
        GRM (Untrained) + RL & 53.8 & 63.1 & 3.1 & 79.1 & 70.0 \\
        \midrule
        Trained GRM + RL & 69.28 & 70 & 0 & 85.15 & 72.14 \\
        \bottomrule
    \end{tabular}
    }
    \caption{Performance comparison of different models in dialogue quality and solution effectiveness. All values are reported as percentages (\%). The columns represent: Full Score Rate (FSR), Usability Rate (UR), Bottom-line Issue (BI), Accuracy (Acc), and Solution Rate (SR).}
    \label{tab:model_comparison_grm_table8}
\end{table}

\begin{figure}[h]
\centering
\includegraphics[width=0.8\textwidth]{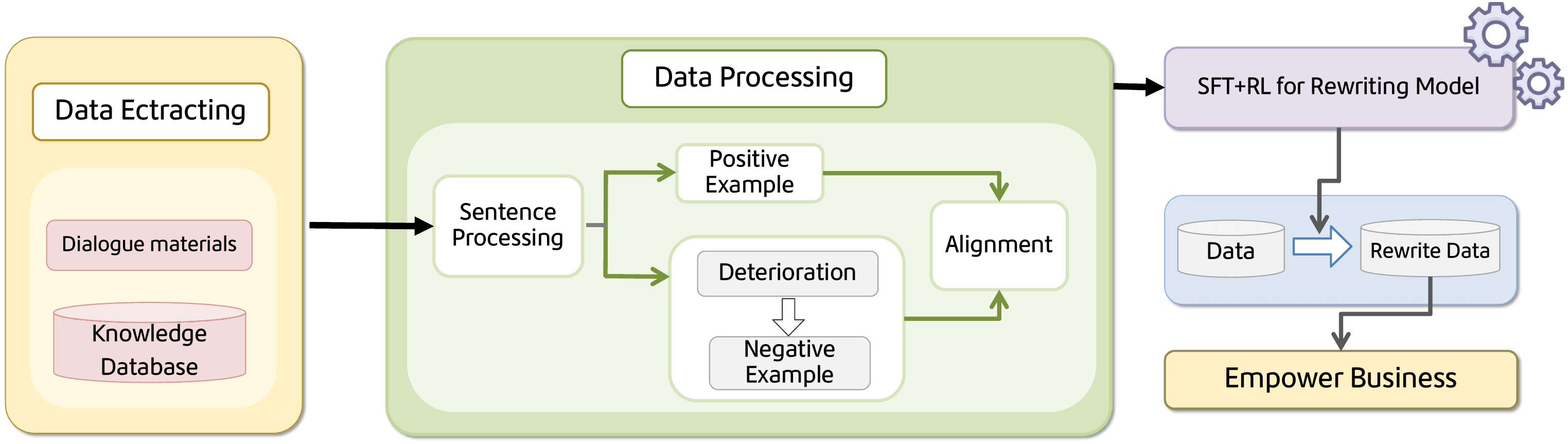}
\caption{Framework of the solution for enhancing intelligent interaction response quality.}
\label{fig:response_quality_framework}
\end{figure}

\paragraph{Zero-Shot High-Quality Response Rewriting for Intelligent Interaction Conversations.}
In the current telephone intelligent interaction business, the quality of intelligent interaction responses is relatively low. 
Although they meet basic correctness requirements, users are generally dissatisfied, lacking sufficient affirmation and reassurance. 
Existing methods of improving quality through annotation by trainers still face issues of high subjectivity and difficulty in consistently producing high-quality responses. 
We propose a holistic solution to enhance the quality of responses in telephone intelligent interaction, addressing the inadequacies in current practices, which is illustrated in Figure~\ref{fig:response_quality_framework}. 

\textbf{Quality Response Data Acquisition and Evaluation:} This section involves the mining and evaluation of high-quality response data along with degradation and style transfer techniques. By presetting response evaluation dimensions, we mine high-quality dialogue corpora, selecting responses with multi-dimensional labels (at least two dimensions) as training samples. During this stage, precision and recall are computed through manually annotated test sets. Evaluation of rewritten responses includes win rate, plan consistency rate, and the proportion of Good/Bad/Same, supported by both automated and manual assessments.

Additional evaluation metrics include:

\begin{itemize}
    \item \textit{Win Rate:} This measures the success of rewritten responses compared to the original ones: $$ W = \frac{\sum \mathbb{I}(\text{Rewritten response} > \text{Original response})}{N}$$
    \item \textit{Plan Consistency Rate:} Compares the consistency of solutions before and after rewriting: $$ C = \frac{\sum (\text{\# Consistent Solutions})}{\sum (\text{\# Consistent Solutions}) + \sum (\text{\# Inconsistent Solutions})}.$$
\end{itemize}
where $\mathbb{I}$ is the indicator function, and $N$ denotes the testset size. A higher win rate indicates that the rewritten model's responses are more frequently preferred over the original model's outputs, suggesting that the rewriting strategy leads to improved overall model performance. Therefore, if the win rate of the rewritten responses is significantly higher, it demonstrates the superiority and effectiveness of the rewriting approach in enhancing response quality.

These metrics can be obtained through both automated evaluation and manual annotation. This process also involves degradation, rewriting, and style transfer. Degradation rewriting uses large models or real online models to transform quality responses into degraded versions, characterized by mechanization, singular reassurance, or ambiguous plans as comparison samples. Style transfer introduces a dedicated response style conversion module, ensuring rewritten responses maintain the original solution while aligning style and emotional expressions closely with quality standards.

\textbf{Model Training and Reward Mechanism Design:} Adopt a two-stage training process of SFT+RL. The reward function based on semantic similarity (BGE embedding) penalizes repetitive and overly similar responses, adjusting the similarity threshold to ensure diversity and consistency. Removing the reward mapping mechanism enhances the model's ability to learn subtle differences in responses.

Additional aspects of the reward function design include:
\begin{itemize}
    \item Redefining similarity without reward mapping: Utilizing the BGE embedding model to calculate semantic similarity between the generated response and the ground truth.
    \item Penalty for duplicate responses: Employing Jaccard calculation within the current dialogue, penalizing duplication by half for any given dialogue turn if its similarity exceeds a threshold.
    \item Reward structure for response similarity: During RFT training, construct an embedding of historical responses for comparison against current responses in the training dataset, maintaining the top 5 with similarity exceeding the threshold. 
\end{itemize}

These mechanisms collectively aim to refine the learning process, ensuring nuanced adjustments of response generation to enhance overall quality.

Key indicators in Table~\ref{tab:performance_comparison} show significant improvements: Fulfillment resolution rate increased by 11.0 percentage points (50.4\%→61.4\%), after-sales resolution rate increased by 16.4 percentage points (38.1\%→54.5\%), and overall scenario resolution rate improved by 5.4 percentage points (46.6\%→52.0\%). Rewritten responses show significant improvements in personalization and colloquial expression, closer to human intelligent interaction impressions. Typical scenarios, such as guiding user operations, explaining refund amounts, rider delivery anomalies, and invoice applications, exhibit stronger emotional value and problem-solving capabilities.

\begin{table}[h!]
\centering
{\small
\begin{tabular}{@{}lccc@{}}
\toprule
\textbf{Scenario} & \textbf{Service Volume} & \textbf{Queue Rate} & \textbf{Experiences} \\ 
\midrule
Overall (E) & 13398 & 34.6$^{+4.6}$ & 52.0$^{+5.4}$ / 60.7$^{+3.2}$ \\ 
Overall (C) & 13288 & 30.0 & 46.6 / 57.5 \\ 
\midrule
Fulfillment (E) & 3589 & 24.9$^{+3.5}$ & 61.4$^{+11.0}$ / 61.4$^{+11.0}$ \\ 
Fulfillment (C) & 2301 & 21.4 & 50.4 / 50.4 \\ 
\midrule
After-Sales (E) & 1603 & 29.0$^{+5.3}$ & 54.5$^{+16.4}$ / 64.7$^{+17.1}$ \\ 
After-Sales (C) & 1495 & 23.7 & 38.1 / 47.6 \\ 
\bottomrule
\end{tabular}
}
\caption{Performance across different scenarios, where (E) and (C) represent the Experimental and Control Group, respectively. Queue Rate and Experiences values are reported as percentages (\%).}
\label{tab:performance_comparison}
\end{table}

This solution improves interaction quality and resolution rates via zero-annotation response rewriting, combining large model optimization with automated evaluation for human-like performance.

\section{Agent Section}
\label{sec:multi_agent}
In many complex scenarios, a single agent often fails to meet the diverse requirements.
Multi-Agent Systems~(MASs), by enabling collaboration and coordination among specialized agents, substantially amplify the capabilities of our individual LLMs.
Their strength in knowledge memorization allows distributed agents to maintain and exchange heterogeneous knowledge bases while preventing overload on any individual system~\citep{tran2025multiagentcollaborationmechanismssurvey,Hatalis_Christou_Myers_Jones_Lambert_Amos-Binks_Dannenhauer_Dannenhauer_2024,10.1145/3748302}. We propose a multi-agent architecture in which specialized sub-agents handle distinct functions and emit signals that explicitly influence the master agent. 
This design aligns with the Agents-as-Tools framework~\citep{openai-agents-tools}, a hierarchical collaborative architecture where the primary agent maintains conversational control while dynamically invoking specialized agents to address subtasks. 
Distinct from complete conversational handover, this paradigm treats the specialized agents as executable tools: the primary agent invokes them to get information or handle part of a request, then incorporates their results back into its own response to the user.
This approach demonstrates adaptive agility, enabling real-time determination of sub-agent activation based on query requirements rather than predefined workflows.
For example, during an interaction between the primary conversational agent and a user, the system may encounter situations of information scarcity or produce uncertain or unjustified decisions. In such cases, a corresponding sub-agent is dynamically invoked to perform the required subtask. Meanwhile, the primary agent remains in continuous communication with the user, managing the dialogue flow and awaiting the sub-agent’s response, thereby ensuring a coherent and seamless interactive experience.

Additionally, we incorporate elements of the Handoff pattern~\citep{openai-agents-handoffs}, a decentralized multi-agent orchestration in which an agent can delegate a task to another specialized agent, transferring control and context in a one-way handoff. 
 While this pattern offers limited flexibility and incurs high development overhead, its transparent workflow guarantees high reliability.
 By combining the adaptive agility of Agents-as-Tools with the reliability of the Handoff pattern, our system ensures that the master agent ultimately decides adoption of signals based on real-time signals and the dialogue context. If adopted, these signals are executed only at appropriate moments. This hybrid design enables the system to interact naturally while delivering a high-quality communication experience.

\subsection{Outbound-Call Agent}

Intelligent outbound calling is implemented in four sequential steps. The master dialogue agent first emits a valid outbound request. A parsing sub-agent then extracts the call parameters, an execution sub-agent places the call, and a collection sub-agent retrieves the outcome. The master agent finally consolidates the information returned by the sub-agents and reports the result to the user.

To assess the effectiveness of the outbound-call agent, we compare the scores before and after its introduction along three dimensions: the correctness of the information conveyed, adherence to the outbound procedure, and the accuracy of user intent understanding. Averaging across these dimensions yields the final results reported in Table~\ref{tab:performance comparison outbound}.

\begin{table}[htbp]
    \centering
    {\footnotesize
    \begin{tabular}{lcc}
        \toprule
        \textbf{Score} & \textbf{Baseline} & \textbf{Outbound-Call Agent} \\
        \midrule
        Average Score & $57$ & $80^{+23}$ \\
        \bottomrule
    \end{tabular}
    }
    \caption{Performance comparison before and after introducing the outbound-call agent. The upper part shows the relative accuracy improvement over the baseline, while the lower part reports the absolute accuracy values.}
    \label{tab:performance comparison outbound}
\end{table}

\subsection{Agent of Proactive Collaboration}
Intelligent interaction systems in real-world business environments are designed to support multiple scenarios concurrently. Thus, in many user-initiated interactions, users are generally required to first supply standardized input—whether by describing their issue or selecting from predefined choices via interactive cards—before the system can direct the request to the relevant sub-scenario and model for further handling.
Such a system architecture, however, depends on multiple sub-scenario models and lacks the flexibility to switch across scenarios. Furthermore, users are required to actively complete standardized queries or option selections, while the intelligent interaction system merely responds passively. This results in lower communication efficiency, diminished user willingness to engage with the system, and ultimately hinders effective problem resolution.

\begin{figure}[t]
\centering
\includegraphics[width=0.8\textwidth]{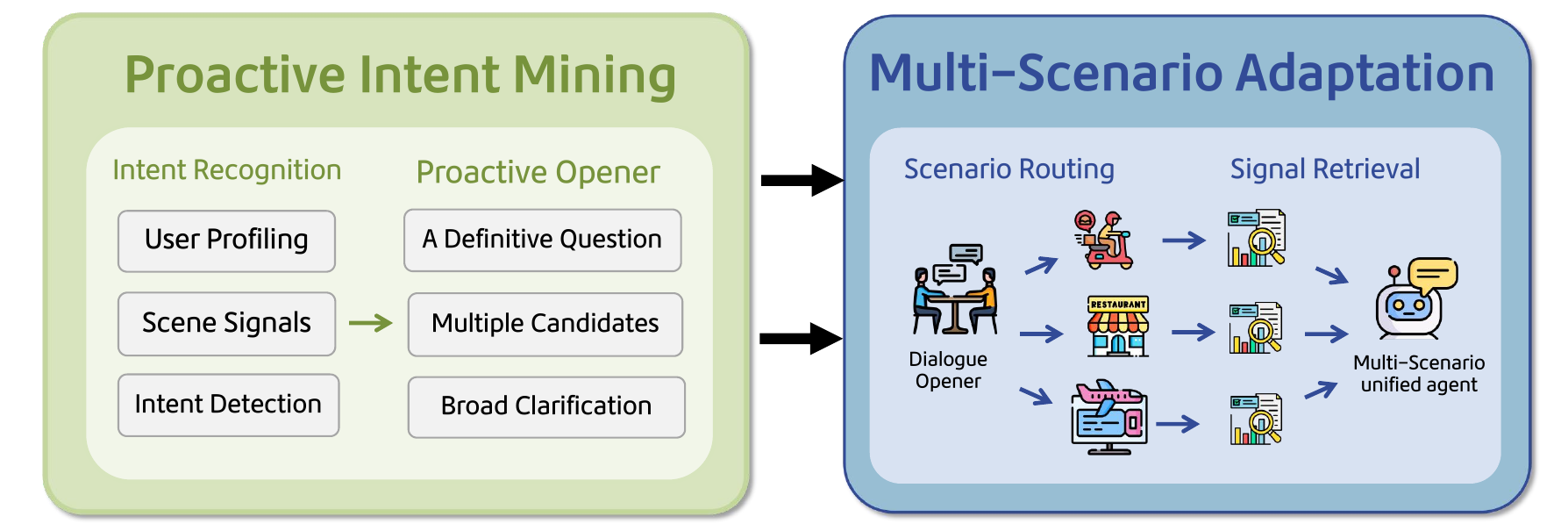}
\caption{Framework of the proactive collaboration mechanism.}
\label{fig:active_collaboration}
\end{figure}

To overcome these limitations, we incorporate an agent component of proactive collaboration that works across all service scenarios, which can be further decoupled into two strategies: proactive intent mining at the start of a conversation and multi-scenario adaptation. As shown in Figure~\ref{fig:active_collaboration}, to improve user engagement, the proactive intent mining module detects potential user needs and confirms them proactively. The system first checks whether the initial signal is strong enough to support an accurate prediction. If it is, the system proposes a single likely issue and generates a short three-part script: explain the signal used, confirm the issue with the user, and then either proceed or provide a solution. If the signal isn’t strong enough, the system offers a list of possible issues for the user to pick from. As shown in Table~\ref{tab:performance_comparison_of_active_service}, by taking the initiative in this way, the system makes it easier for users to stay engaged and more willing to keep the conversation going.

\begin{table}[h!]
\centering
{\footnotesize
\begin{tabular}{@{}lcc@{}}
\toprule
\textbf{Model} & \textbf{USM 1 $\downarrow$} & \textbf{USM 2 $\uparrow$} \\
\midrule
Baseline & 18.2 & 48.0 \\ 
Baseline with Proactive Collaboration  & 12.5  & 58.8 \\ 
Improvement $\Delta$ & -5.6 & +10.8 \\
\bottomrule
\end{tabular}
}
\caption{Performance comparison (\%) with and without proactive collaboration strategy for the User Satisfaction Metric 1 (USM 1) and User Satisfaction Metric 2 (USM 2).}
\label{tab:performance_comparison_of_active_service}
\end{table}

After confirming user requirements, the multi-scenario adaptation strategy identifies the appropriate service scenario based on the user’s reply. By leveraging predefined rules, the system retrieves the relevant signals and operational instructions for each scenario, enabling seamless switching between scenarios within a single dialogue while maintaining workflow consistency. Notably, we adopt a unified interaction service agent, which is trained as described in previous sections, rather than separate models for each scenario. Specifically, this all-scenario-unified model is trained on varied contexts containing redundant signals, and during deployment, it only retrieves the relevant signals for inference. This design reduces the training and deployment burden of maintaining multiple models while also improving the user experience in multi-scenario dialogues.

\subsection{Agent of Multi-Modal Understanding}
\begin{figure}[t]
\centering
\includegraphics[width=0.6\textwidth]{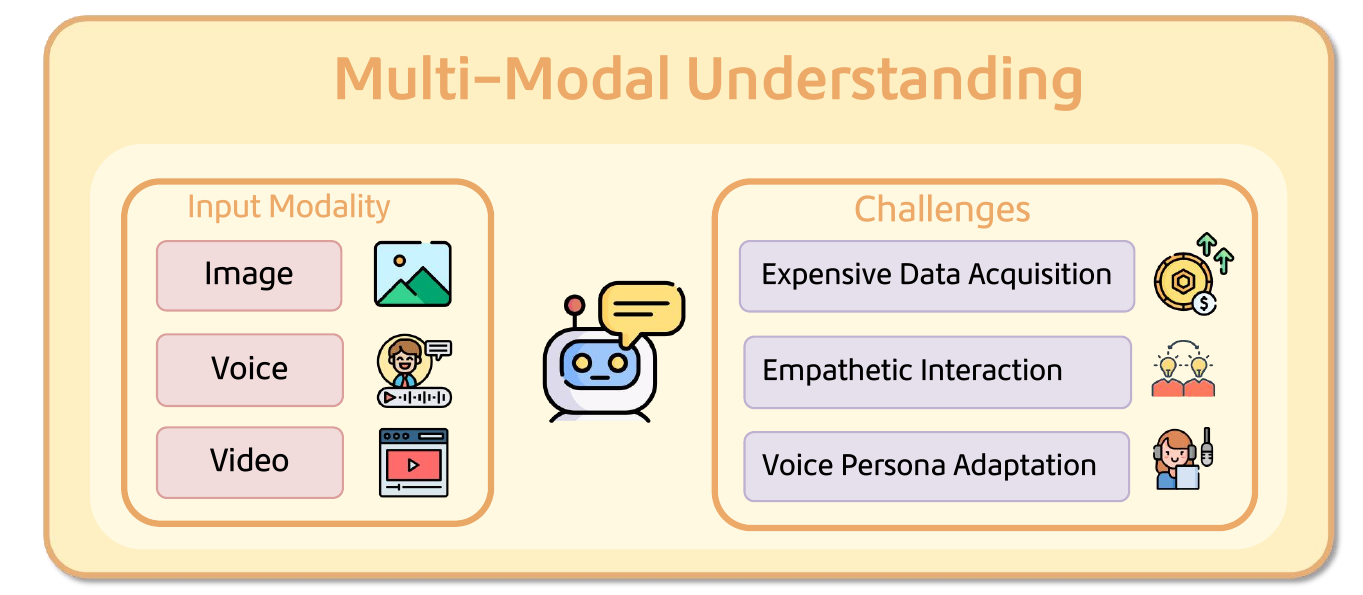}
\caption{Challenges of multi-modal understanding.}
\label{fig:mm_understanding}
\end{figure}
Understanding multimodal user inputs~\citep{DBLP:conf/acl/CaffagniCBMS0CC24} is another key agentic capability for intelligent interaction systems.  In practical deployment, a considerable fraction of dialogue interactions involves non-textual modalities, including speech, images, and videos. Within the agent workflow, accurately recognizing such inputs is critical for planning the next steps. Therefore, within the end-to-end service pipeline, we integrate a multimodal agent for image recognition to assist the primary dialogue model, as shown in Figure~\ref{fig:mm_understanding}. During the exploration phase, we first carried out rapid validation with open-source multimodal models. After confirming their positive impact on online business metrics, we began training a business-oriented multimodal model. The image recognition capability was then expanded to cover all interaction service scenarios. Ultimately, our self-developed model not only achieved higher accuracy than the open-source GPT-4o~\citep{achiam2023gpt} on image perception tasks but also delivered a marked reduction in application latency. Beyond visual inputs, speech plays a central role in interaction services. To advance this capability, we have pursued extensive work on end-to-end speech large models. The resulting multimodal speech systems have achieved several technical milestones, including tighter multimodal alignment, seamless integration of function-calling, and enhanced safety mechanisms.

\section{Evaluation Section}
\begin{figure}[t]
\centering
\includegraphics[width=\textwidth]{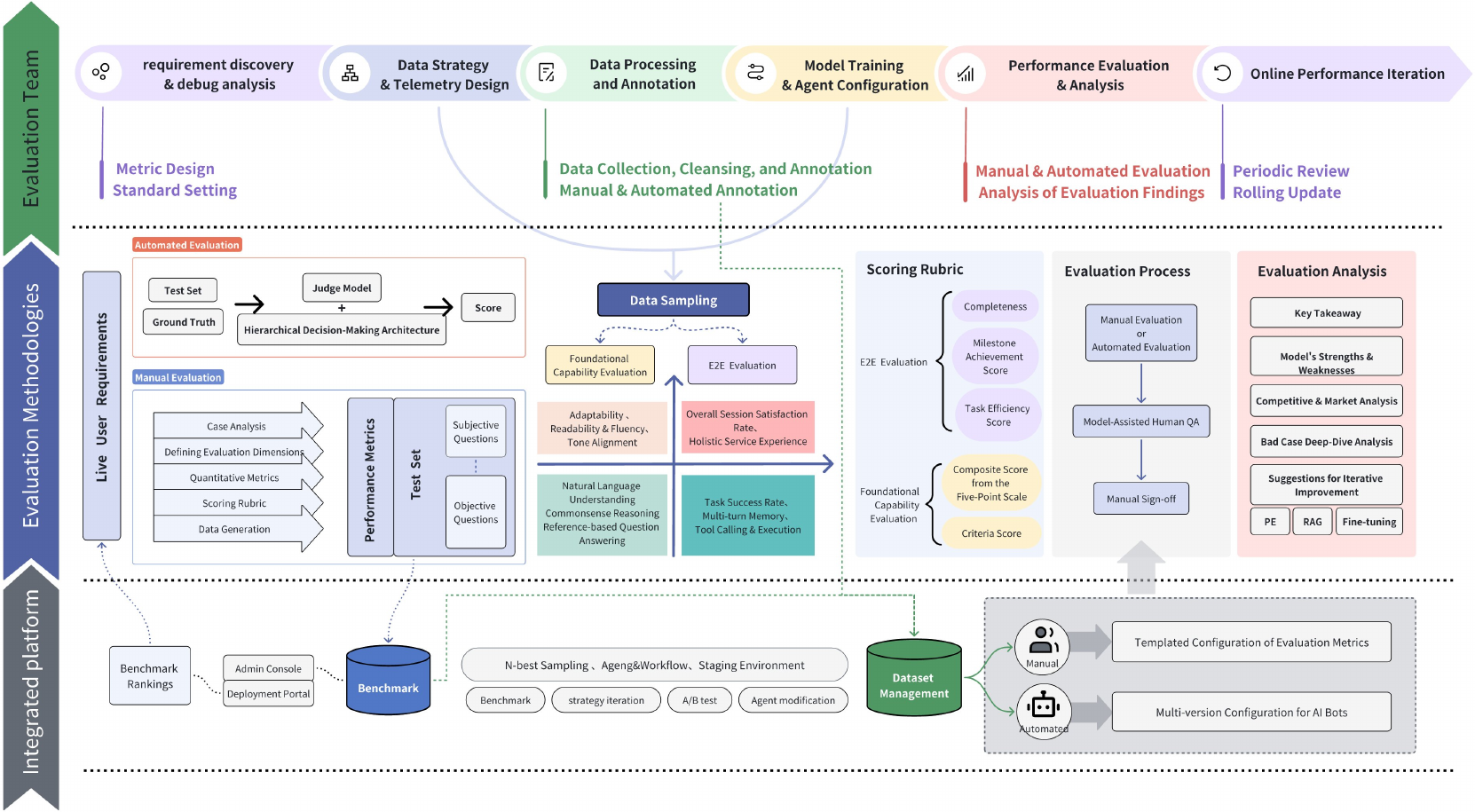}
\caption{Framework of the evaluation platform for models in Intelligent Interaction Systems.}
\label{fig:evaluation_framework}
\end{figure}

To address the lack of a standardized benchmark in Intelligent Interaction Systems, we propose an evaluation framework. This framework is a unified platform supporting an end-to-end workflow with both manual and automated evaluation (Section \ref{sec:automated_eval}).
As illustrated in Figure \ref{fig:evaluation_framework}, the platform integrates functionalities like corpus management, version control, and process monitoring. We use it to periodically evaluate SOTA models (Section \ref{sec:base_model_eval}) and AI agents (Section \ref{sec:agent_eval}). The results are published on a dynamic leaderboard to guide algorithmic development.
Our framework consists of two components: 1) Evaluation Set Construction and 2) Evaluation Execution.

\paragraph{Evaluation Set Construction.}
Our evaluation set is constructed in three stages:
\textbf{1) Data Distribution Design.} We design data distributions based on real-world scenarios. The dataset is periodically refreshed to reflect evolving conditions. \textbf{2) Data Sampling.} We use stratified random sampling based on key dimensions like difficulty and length. Other dimensions follow normal or uniform distributions to create a representative data sample. \textbf{3) Metric Design.} Metrics are designed based on the scenario. We use established metrics, including GSB, Mean Opinion Score (MOS), and Perfect Rates.

\paragraph{Evaluation Execution.}
Evaluation includes two types: human and model-based.
\textbf{1) Human Evaluation.} Human evaluation uses a blind, multi-assessor protocol. To ensure reliability, each case is independently evaluated by two to three assessors who are unaware of the model's identity.
\textbf{2) Model-based Evaluation.} Model-based evaluation uses a dedicated model for automated scoring and rationale generation. Automated results are cross-validated with human annotations, and human experts resolve any disagreements. We analyze scoring errors to iteratively refine the evaluation model, targeting a human-machine agreement rate of over 95\%.

\subsection{Model-based Automated Evaluation}
\label{sec:automated_eval}

Manual evaluation is costly and inconsistent. To address this, we developed an automated evaluation framework that assigns a four-tier score to single-turn responses: -1 for ``red-line'' behaviors (e.g., rule violations), and 0/1/2 for unsatisfactory, satisfactory, and excellent performance, respectively.

\paragraph{Evaluation Framework.}
We propose a four-stage pipeline architecture combining multi-agent collaboration and hierarchical decision-making to simulate human expert evaluation.

\textbf{Stage 1: Context Aggregation and Input Preparation.} This stage prepares structured inputs, including the [Historical Dialogue], [Model's Response], a Golden Standard answer with its CoT, and Scoring Rubrics.

\textbf{Stage 2: Parallelized Scoring by Specialized Agents.} This stage decomposes the task into three parallel binary classifications, each handled by a specialized agent: the -1 Point Adjudicator for ``red-line'' behaviors, the 2 Point Adjudicator for perfect responses, and the 0-Point Adjudicator for ineffective responses. The score of `1' is assigned when none of these conditions are met.

\textbf{Stage 3: Efficient Decision Routing.} This stage aggregates results from the parallel agents. It resolves non-conflicting cases directly and routes conflicts to the next stage to improve efficiency.

\textbf{Stage 4: Expert-Level Conflict Adjudication.} This final stage uses an ``Expert Adjudicator'' model with deeper reasoning capabilities to resolve ambiguous cases and make the final decision.

\subsection{Base Model Evaluation}
\label{sec:base_model_eval}

We evaluate base models in isolation, decoupled from business systems, external tools, and knowledge bases. For this purpose, we built the Intelligent Interaction Systems Benchmark, which uses a capability taxonomy derived from real-world scenarios.

\begin{figure}[t]
\centering
\includegraphics[width=\textwidth]{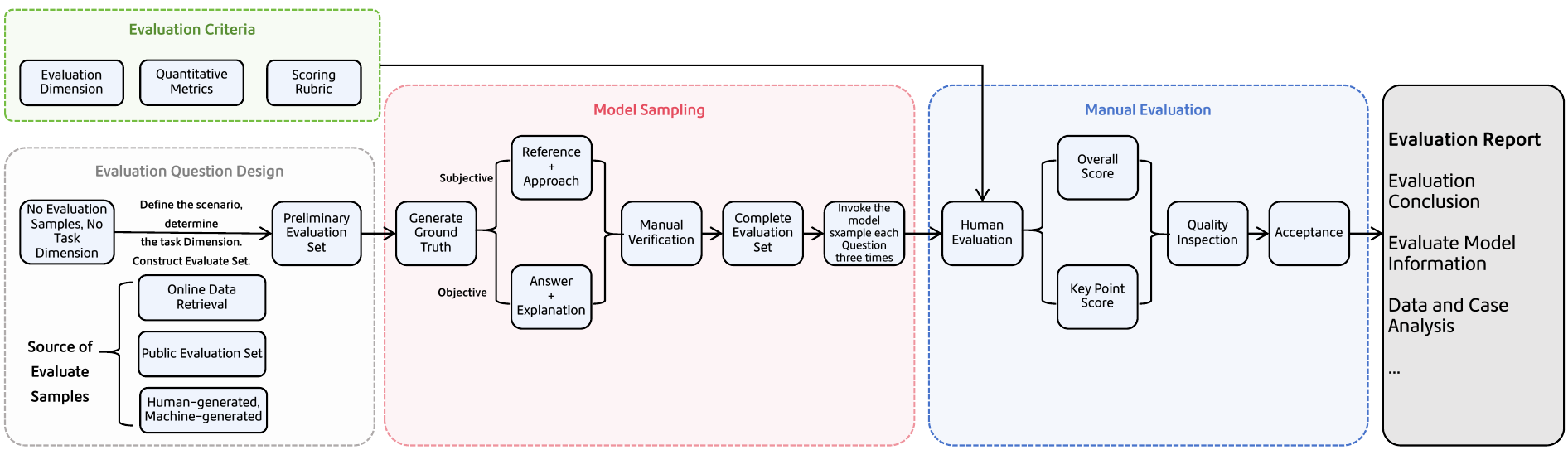}
\caption{Framework of the evaluation for base models.}
\label{fig:base_model_eval}
\end{figure}

The evaluation focuses on the model's intrinsic capabilities—such as problem comprehension, communication, and resolution—rather than its end-to-end performance in a production environment. These metrics are used to inform business decisions and guide algorithm optimization. The evaluation framework is illustrated in Figure \ref{fig:base_model_eval}.

\begin{table}
\centering
\caption{Performance of multiple models in the Intelligent Interaction Systems Benchmark. Green indicates the top three highest scores, while red indicates the bottom three lowest scores.}
\label{tab:4_evaluation}
\resizebox{\textwidth}{!}{

\begin{tblr}{
  row{1,2} = {c},
  colspec = {l *{12}{c}},
  cell{1}{1} = {r=2}{},
  cell{1}{2} = {r=2}{},
  cell{1}{3} = {r=2}{},
  cell{1}{4} = {r=2}{},
  cell{1}{5} = {r=2}{},
  cell{1}{6} = {r=2}{},
  cell{1}{7} = {c=2}{},
  cell{1}{9} = {c=3}{},
  cell{1}{12} = {c=2}{},
  vline{2-7,9,12} = {1-2}{},
  vline{8,10,11,13} = {2}{},
  vline{2-13} = {3-15}{},
  hline{1,3-16} = {-}{},
  hline{2} = {7-13}{},
  cell{3}{2} = {bg=GreenHaze, c},
  cell{4}{2} = {bg=Jade, c},
  cell{5}{2} = {bg=Shamrock, c},
  cell{12}{2} = {bg=Salmon, c},
  cell{13}{2} = {bg=SunsetOrange, c},
  cell{14}{2} = {bg=AlizarinCrimson, c},
  cell{3}{3} = {bg=GreenHaze, c},
  cell{4}{3} = {bg=Jade, c},
  cell{5}{3} = {bg=Shamrock, c},
  cell{13}{3} = {bg=Salmon, c},
  cell{12}{3} = {bg=SunsetOrange, c},
  cell{14}{3} = {bg=AlizarinCrimson, c},
  cell{4}{4} = {bg=GreenHaze, c},
  cell{5}{4} = {bg=Jade, c},
  cell{3}{4} = {bg=Shamrock, c},
  cell{12}{4} = {bg=Salmon, c},
  cell{13}{4} = {bg=SunsetOrange, c},
  cell{11}{4} = {bg=AlizarinCrimson, c},
  cell{14}{5} = {bg=GreenHaze, c},
  cell{10}{5} = {bg=Jade, c},
  cell{8}{5} = {bg=Shamrock, c},
  cell{9}{5} = {bg=Salmon, c},
  cell{12}{5} = {bg=SunsetOrange, c},
  cell{11}{5} = {bg=AlizarinCrimson, c},
  cell{3}{6} = {bg=GreenHaze, c},
  cell{6}{6} = {bg=Jade, c},
  cell{5}{6} = {bg=Shamrock, c},
  cell{14}{6} = {bg=Salmon, c},
  cell{12}{6} = {bg=SunsetOrange, c},
  cell{13}{6} = {bg=AlizarinCrimson, c},
  cell{9}{7} = {bg=GreenHaze, c},
  cell{3}{7} = {bg=Jade, c},
  cell{7}{7} = {bg=Shamrock, c},
  cell{12}{7} = {bg=Salmon, c},
  cell{5}{7} = {bg=SunsetOrange, c},
  cell{11}{7} = {bg=AlizarinCrimson, c},
  cell{4}{8} = {bg=GreenHaze, c},
  cell{3}{8} = {bg=Jade, c},
  cell{6}{8} = {bg=Shamrock, c},
  cell{9}{8} = {bg=Salmon, c},
  cell{12}{8} = {bg=SunsetOrange, c},
  cell{13}{8} = {bg=AlizarinCrimson, c},
  cell{3}{9} = {bg=GreenHaze, c},
  cell{5}{9} = {bg=Jade, c},
  cell{6}{9} = {bg=Shamrock, c},
  cell{12}{9} = {bg=Salmon, c},
  cell{13}{9} = {bg=SunsetOrange, c},
  cell{14}{9} = {bg=AlizarinCrimson, c},
  cell{3}{10} = {bg=GreenHaze, c},
  cell{6}{10} = {bg=Jade, c},
  cell{5}{10} = {bg=Shamrock, c},
  cell{13}{10} = {bg=Salmon, c},
  cell{12}{10} = {bg=SunsetOrange, c},
  cell{14}{10} = {bg=AlizarinCrimson, c},
  cell{3}{11} = {bg=GreenHaze, c},
  cell{5}{11} = {bg=Jade, c},
  cell{4}{11} = {bg=Shamrock, c},
  cell{13}{11} = {bg=Salmon, c},
  cell{12}{11} = {bg=SunsetOrange, c},
  cell{14}{11} = {bg=AlizarinCrimson, c},
  cell{5}{12} = {bg=GreenHaze, c},
  cell{4}{12} = {bg=Jade, c},
  cell{3}{12} = {bg=Shamrock, c},
  cell{12}{12} = {bg=Salmon, c},
  cell{14}{12} = {bg=SunsetOrange, c},
  cell{13}{12} = {bg=AlizarinCrimson, c},
  cell{4}{13} = {bg=GreenHaze, c},
  cell{5}{13} = {bg=Jade, c},
  cell{11}{13} = {bg=Shamrock, c},
  cell{12}{13} = {bg=Salmon, c},
  cell{13}{13} = {bg=SunsetOrange, c},
  cell{14}{13} = {bg=AlizarinCrimson, c},
}
\textbf{Model} & {\textbf{Overall }\\\textbf{Score}} & {\textbf{Availability }\\\textbf{Rate}} & {\textbf{Perfect }\\\textbf{Rate}} & {\textbf{Problem}\\\textbf{communicate}} & {\textbf{Solution}\\\textbf{ Framework}} & \textbf{Problem Understanding} & & \textbf{Detail Accuracy} & & & \textbf{Expression Style} & \\
& & & & & & {\textbf{Semantic}\\\textbf{ Understand}} & {\textbf{Knowledge}\\\textbf{ Understand}} & {\textbf{Truth-}\\\textbf{fulness}} & {\textbf{Comprehen-}\\\textbf{siveness}} & {\textbf{Instruction }\\\textbf{Following}} & {\textbf{Human-}\\\textbf{like}} & {\textbf{Persona-}\\\textbf{lization}} \\
Doubao-Seed-1.6 & 339.33 & 45.24\% & 14.40\% & 418 & 316.67 & 541.67 & 520.67 & 300 & 291.33 & 287.33 & 168.33 & 172.67 \\
GPT-oss-120B & 321.44 & 38.44\% & 22.22\% & 403.11~ & 267.21~ & 536.39~ & 529.20~ & 264.60~ & 249.57~ & 261.99~ & 177.05~ & 233.24~ \\
LongCat & 305.11 & 34.22\% & 15.33\% & 403.11~ & 269.17~ & 492.61 & 471.05~ & 272.44~ & 258.72~ & 272.44~ & 182.93~ & 218.21~ \\
GPT-4.1 & 270.67 & 29.59\% & 12.47\% & 424.67 & 281.67 & 524.33 & 510.33 & 266.33 & 260.33 & 251.33 & 145 & 152 \\
Qwen-Max & 260.67 & 29.82\% & 10.20\% & 430.33 & 268.67 & 540 & 496.67 & 259 & 244.67 & 256.33 & 148 & 152.33 \\
GPT-4o-2024-11-20 & 249 & 29.59\% & 9.30\% & 445.33 & 260.67 & 515.33 & 492 & 258 & 247 & 234 & 135.33 & 135 \\
Claude-Sonnet-4 & 247.67 & 31.29\% & 12.36\% & 398.67 & 235 & 544.33 & 434 & 229 & 218.67 & 222.67 & 147 & 152.33 \\
Deepseek-V3-0324 & 236 & 29.37\% & 10.88\% & 460.33 & 228 & 539.67 & 453.67 & 229 & 216.33 & 210.33 & 129.33 & 134.67 \\
Claude-Opus-4.1 & 207.76 & 22\% & 4.00\% & 337.12~ & 234.55~ & 491.31~ & 452.11~ & 231.28~ & 211.68~ & 222.13~ & 132.63~ & 188.81~ \\
Qwen-Turbo & 175.33 & 21.32\% & 7.48\% & 368.33 & 174.33 & 506.33 & 416.67 & 163.33 & 147.67 & 155.33 & 94 & 99.33 \\
DeepSeek-R1-0528 & 171.67 & 21.77\% & 6.80\% & 428 & 159.67 & 539 & 306 & 157 & 153 & 159 & 71 & 81 \\
Qwen3-32B & 162.67 & 20.63\% & 7.48\% & 466.33 & 184.67 & 529.33 & 500 & 144.33 & 139.67 & 142.33 & 78.67 & 79 \\
\end{tblr}
}
\end{table}

\paragraph{Evaluation Method and Dataset.} The dataset is primarily composed of cleaned and processed data retrieved from live online traffic, supplemented by a selection of manually constructed data. We employ a stratified sampling strategy based on specific ratios for dimensions such as difficulty and length, while other dimensions are randomly sampled following a normal or uniform distribution. Each item in the evaluation set contains the necessary input prompt, metadata tags (e.g., key assessment points), and the ground truth for comparison with the model's output. The final score for each item is the average of three separate runs. The final results are shown in Table~\ref{tab:4_evaluation}, where \textbf{Overall Score} means the sum of scores from all test cases in the evaluation set (scoring range: -1 to 3 points), \textbf{Availability Rate} means the percentage of test cases in the evaluation set that scored 2 or 3 points, measuring the model's ability to generate available answers, \textbf{Perfect Rate} means the proportion of test cases that scored 3 points in the evaluation set, indicating the model's capability to generate highly effective answers.
We can find that the top tier in the Intelligent Interaction Systems includes Doubao-Seed-1.6, GPT-oss-120B, and LongCat.

\subsection{End-to-End Agent Evaluation}
\label{sec:agent_eval}

Intelligent agents in current interaction systems typically employ a workflow that integrates LLMs with external tools and knowledge bases. This process involves retrieving context, making decisions, and executing actions, often stabilized by safeguards like circuit breaking. The architecture's reliance on fixed paths ensures high consistency for well-defined tasks but lowers the demand for the model's autonomous planning and exploration capabilities.

Consequently, a comprehensive end-to-end evaluation must adopt a dual perspective. It must assess the agent's overall task completion efficiency while simultaneously tracing its internal trajectory to conduct a fine-grained evaluation of its execution logic and key decisions.

\begin{figure}[t]
\centering
\includegraphics[width=\textwidth]{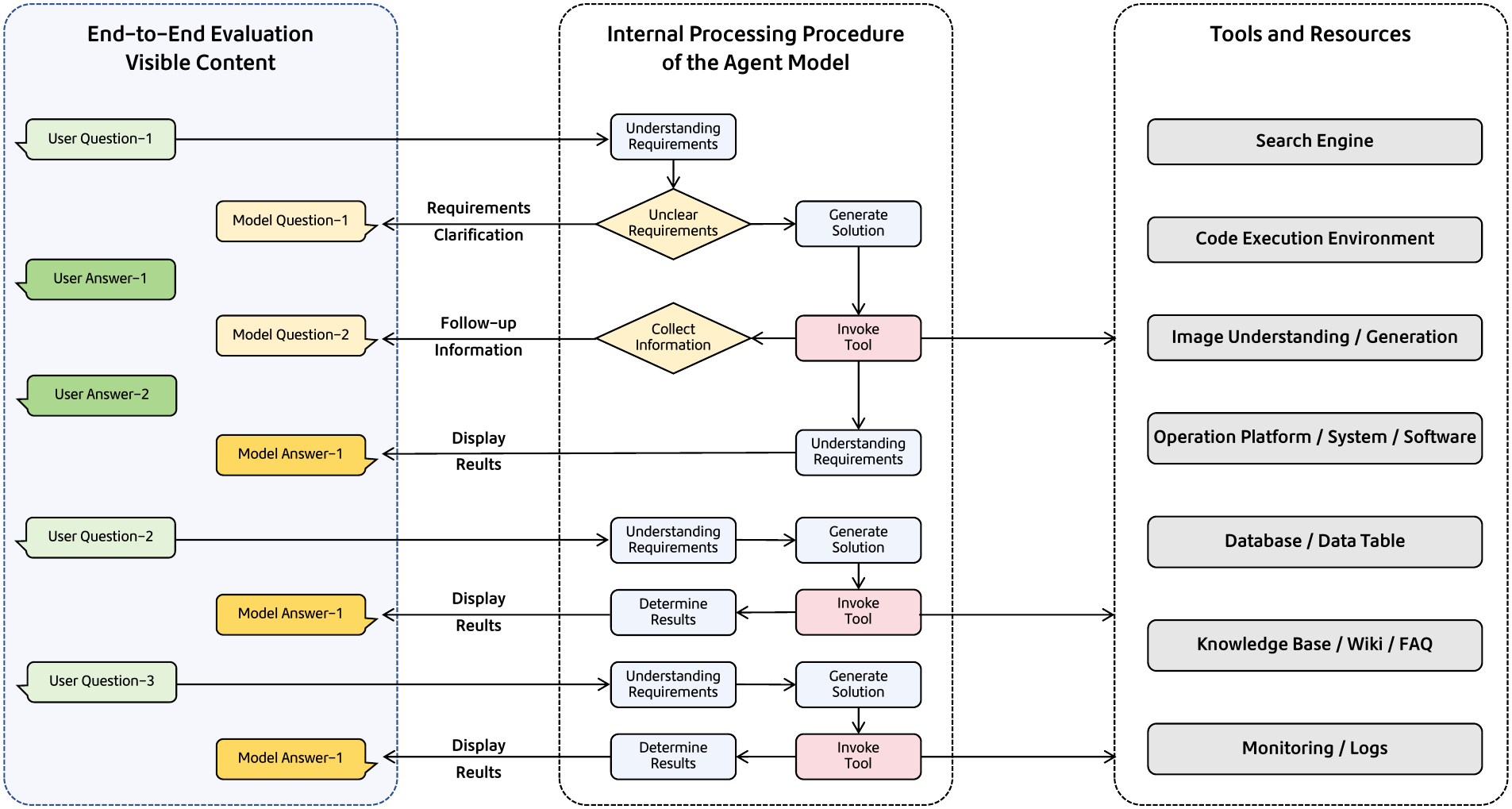}
\caption{Framework of the evaluation for agents.}
\label{fig:agent_eval}
\end{figure}

\paragraph{Evaluation Framework Design and Methodology.}
This framework evaluates the agent's end-to-end effectiveness (Figure \ref{fig:agent_eval}), covering the core service chain while excluding UI elements. The evaluated agents use a modular, multi-agent design where a Lead Agent decomposes tasks, coordinates Sub-agents, and integrates their final results. This architecture excels at complex, parallel tasks with long contexts. A key challenge is creating a realistic sandbox environment with a complete simulated ecosystem, including databases, user simulators, APIs, and business rules.

Within this framework, the evaluation not only assesses whether the Agent resolves the user's problem from end to end but also scrutinizes whether its performance at Key Checkpoints meets expectations. Furthermore, it holistically considers cost, efficiency, security, and the integration with upstream and downstream ticketing systems.

\section{Conclusion}
\label{sec:Conclusion}
This report systematically reviews the practical implementation, key challenges, and future directions of generative artificial intelligence in the field of Intelligent Interaction Systems. Based on real business requirements, we propose a technical solution centered on multi-stage training and multi-agent architecture, and thoroughly analyze its distinctive advantages and potential impact on the future development of intelligent interaction systems.

With the rapid expansion of enterprise business, there is an increasing demand for consistent user experience, operational standardization, and efficient, cost-effective service delivery. Generative artificial intelligence, leveraging advanced language understanding and generation capabilities, has emerged as the preferred technology for Intelligent Interaction Systems, enabling simultaneous improvements in both service costs and user experience through Pareto optimization. However, in complex multi-turn dialogue scenarios, Intelligent Interaction Systems face significant challenges: they must exhibit highly human-like interactive abilities, accurately comprehend user inquiries, and strictly adhere to dynamic business rules, all while maintaining adaptability, optimizing effectiveness, and expanding coverage at low cost. To address these multifaceted challenges, we propose a comprehensive technical solution that constructs and iteratively refines high-quality data, utilizes a multi-stage training paradigm, and introduces a robust multi-agent architecture. This innovative approach continuously enhances the model’s capabilities within the business domain, enables proactive and operable service modes, and effectively broadens the scope of service. Compared to existing market products that focus primarily on pre-sales scenarios and tool invocation, our solution targets after-sales service, emphasizing the model’s communication and negotiation abilities. This focus ensures the system’s flexibility, scalability, and effectiveness in solving real user problems. As technology continues to advance, generative artificial intelligence not only improves the efficiency and quality of Intelligent Interaction Systems but also drives the deep integration of intelligent interaction.

Looking ahead, we anticipate that intelligent interaction systems will evolve into genuinely intelligent, proactive, and user-centric assistants, fundamentally transforming the interaction paradigm between users and digital services. Building on our current foundation, we aim to not only enhance the technical capabilities of our solutions but also broaden their application scope and deepen their integration into users’ daily lives. Our future work will focus on the following three key directions:

$\blacklozenge$\;
\textbf{Empowering Tool Use through Agentic Reinforcement Learning:} By leveraging agentic reinforcement learning, our goal is to improve the model's capacity to autonomously discover, build, and invoke a diverse set of tools. This advancement will enable our system to address a wider spectrum of complex life service scenarios, including intelligent scheduling, personalized recommendations, and end-to-end task execution. Through ongoing reinforcement learning-driven optimization, the system will progressively learn to make more effective decisions within dynamic environments, thereby further improving its abilities and adaptability to real-world challenges.

$\blacklozenge$\;
\textbf{Advancing Multi-Agent Collaboration and Multi-Modal Integration:} We intend to further advance multi-agent collaboration mechanisms, facilitating seamless cooperation among specialized agents with diverse areas of expertise. By integrating multi-modal capabilities—including text, speech, image, and potentially video understanding—the system will be equipped to manage a broader range of complex and varied service scenarios. This collaborative and multi-modal framework is expected to significantly extend the application scope of our solution, thereby enabling it to address increasingly sophisticated user requirements across multiple domains.

$\blacklozenge$\;
\textbf{Building Truly Personalized, User-Centric Assistants:} Our ultimate vision is to create AI assistants that stand from the user’s perspective, proactively think ahead, and genuinely solve user problems—not just as a platform’s service extension, but as trusted partners in users’ daily lives. By leveraging user profiling, preference modeling, and context-aware reasoning, we aim to deliver highly personalized services that adapt to individual users’ habits and intentions. This will foster deeper engagement, higher satisfaction, and a more natural, human-like interaction experience.

Through these bold and innovative approaches, we are committed to continuously advancing the boundaries of intelligent interaction within a diverse array of daily service scenarios, thereby developing solutions that are both valuable and empowering to all users.
\clearpage
\phantomsection
\addcontentsline{toc}{section}{Contributions}
\section*{Contributions (Alphabetical Order)}

\begin{flushleft}

\textbf{Project Lead}\quad 

Xuxin Cheng$^{1}$, Ke Zeng$^{1}$

\textbf{Core Contributor}\quad

Zhiquan Cao$^{1}$, Linyi Dai$^{1,2}$, Wenxuan Gao$^{1}$, Fei Han$^{1}$, Ai Jian$^{1,13}$, Feng Hong$^{1}$, Wenxing Hu$^{1}$, Zihe Huang$^{1,7}$, Dejian Kong$^{1,2}$, Jia Leng$^{1}$, Zhuoyuan Liao$^{1,2}$, Pei Liu$^{1}$, Jiaye Lin$^{1}$, Xing Ma$^{1}$, Jingqing Ruan$^{1}$, Jiaxing Song$^{1,6}$, Xiaoyu Tan$^{1}$, Ruixuan Xiao$^{1}$, Wenhui Yu$^{1}$, Wenyu Zhan$^{1}$, Haoxing Zhang$^{1}$, Chao Zhou$^{1,2}$, Hao Zhou$^{1}$, Shaodong Zheng$^{1}$

\textbf{Contributor}\quad

Ruinian Chen$^{1}$, Siyuan Chen$^{1}$, Ziyang Chen$^{1,10}$, Yiwen Dong$^{1}$, Yaoyou Fan$^{1,3}$, Yangyi Fang$^{1,4}$, Yang Gan$^{1,5}$, Shiguang Guo$^{1}$, Qi He$^{1}$, Chaowen Hu$^{1}$, Binghui Li$^{1,12}$, Dailin Li$^{1}$, Xiangyu Li$^{1}$, Yan Li$^{1}$, Chengjian Liu$^{1}$, Xiangfeng Liu$^{1}$, Jiahui Lv$^{1,11}$, Qiao Ma$^{1}$, Jiang Pan$^{1,3}$, Cong Qin$^{1,3}$, Chenxing Sun$^{1}$, Wen Sun$^{1}$, Zhonghui Wang$^{1,9}$, Abudukelimu Wuerkaixi$^{1,4}$, Xin Yang$^{1,6}$, Fangyi Yuan$^{1}$, Yawen Zhu$^{1,5}$, Tianyi Zhai$^{1}$, Jie Zhang$^{1,14}$, Runlai Zhang$^{1,5}$, Yao Xu$^{1,7}$, Yiran Zhao$^{1,2}$, Yifan Wang$^{1,8}$

\textbf{Supervisor}\quad

Xunliang Cai$^{1}$, Yangen Hu$^{1}$, Cao Liu$^{1}$, Lu Pan$^{1}$, Xiaoli Wang$^{1}$, Bo Xiao$^{1}$, Wenyuan Yao$^{1,2}$, Qianlin Zhou$^{1}$, Benchang Zhu$^{1}$

\textbf{Affiliation}

$^{1}$ LongCat Interaction Team, Meituan

$^{2}$ Customer Service and Experience Department, Meituan

$^{3}$ Peking University

$^{4}$ Tsinghua University

$^{5}$ Northeastern University

$^{6}$ Zhejiang University

$^{7}$ University of Chinese Academy of Sciences

$^{8}$ University of Science and Technology of China

$^{9}$ Nanjing University

$^{10}$ National University of Defense Technology

$^{11}$ Beihang University

$^{12}$ Tianjin University

$^{13}$ Beijing University of Posts and Telecommunications

$^{14}$ South China University of Technology

\end{flushleft}
\clearpage

\clearpage
\phantomsection
\addcontentsline{toc}{section}{References}
\bibliographystyle{abbrvnat}
\nobibliography*
\bibliography{bibtex}
\end{document}